\documentclass[conference]{IEEEtran}
\IEEEoverridecommandlockouts
\usepackage{graphicx} 
\usepackage{subcaption}
\usepackage{hyperref}       % hyperlinks

\usepackage{cite}
\usepackage{amsmath,amssymb,amsfonts}
\usepackage{algorithmic}
\usepackage{graphicx}
\usepackage{textcomp}
\usepackage{xcolor}

\def\BibTeX{{\rm B\kern-.05em{\sc i\kern-.025em b}\kern-.08em
    T\kern-.1667em\lower.7ex\hbox{E}\kern-.125emX}}

\begin{document}

\title{Designing an attack-defense game: how to increase the robustness of financial transaction models via a competition}

% \author{\IEEEauthorblockN{ Anonymous authors}}

\author{\IEEEauthorblockN{1\textsuperscript{st} Alexey Zaytsev}
\IEEEauthorblockA{\textit{Skoltech} \\
Moscow, Russia \\
A.Zaytsev@skoltech.ru
}
% \thanks{The work by Alexey Zaytsev was supported by a grant for research centers in the field of artificial intelligence, provided by the Analytical Center for the Government of the Russian Federation in accordance with the subsidy agreement (agreement identifier 000000D730321P5Q0002) and the agreement with the Ivannikov Institute for System Programming of the Russian Academy of Sciences dated November 2, 2021 No. 70-2021-00142.}
\and
\IEEEauthorblockN{2\textsuperscript{nd} Maria Kovaleva}
\IEEEauthorblockA{\textit{Skoltech} \\
Moscow, Russia \\
Maria.Kovaleva@skoltech.ru}
\and
\IEEEauthorblockN{3\textsuperscript{rd} Alex Natekin}
\IEEEauthorblockA{\textit{Open Data Science} \\
Moscow, Russia \\
natekin@ods.ai}
\and
\IEEEauthorblockN{4\textsuperscript{rd} Evgeni Vorsin }
\IEEEauthorblockA{\textit{Innotech} \\
Moscow, Russia \\
vorsineo@gmail.com}
\and
\IEEEauthorblockN{5\textsuperscript{th} Valerii Smirnov}
\IEEEauthorblockA{\textit{Lomonosov Moscow State University} \\
Moscow, Russia\\
smirnovvs@my.msu.ru}
\and
\IEEEauthorblockN{6\textsuperscript{th} Georgii Smirnov}
\IEEEauthorblockA{\textit{Lomonosov Moscow State University} \\
Moscow, Russia \\
Georgii.S.Smirnov@gmail.com}
\and
\IEEEauthorblockN{7\textsuperscript{th} Oleg Sidorshin}
\IEEEauthorblockA{\textit{Kazan Federal University} \\
Kazan, Russia\\
oasidorshin@gmail.com}
\and
\IEEEauthorblockN{8\textsuperscript{th} Alexander Senin}
\IEEEauthorblockA{\textit{Lomonosov Moscow State University} \\
Moscow, Russia \\
aaasenin@gmail.com}
\and
\IEEEauthorblockN{9\textsuperscript{th} Alexander Dudin}
\IEEEauthorblockA{
Kaspersky Lab \\
Moscow, Russia\\
alexander.dudin@outlook.com}
\and
\IEEEauthorblockN{10\textsuperscript{th} Dmitry Berestnev}
\IEEEauthorblockA{\textit{Zvuk} \\
Moscow, Russia \\
Toberest@gmail.com}
}

\maketitle

\begin{abstract}
% Banks routinely use neural networks to make decisions.
% While these models offer higher accuracy, they are susceptible to adversarial attacks, a risk often overlooked in the context of event sequences, particularly sequences of financial transactions, as most works consider computer vision and NLP modalities. 
% Moreover, there is a lack of real-world benchmarks in this area that would imitate the attack-and-defense confrontation.

% We propose a thorough approach to studying these issues: a competition that allows a realistic and detailed investigation of problems in financial transaction data.
% The participants vie with each other, so possible attacks and defenses are examined in close-to-real-life conditions.
% The competition also introduces a unique open dataset featuring financial transactions with credit default labels, enhancing the scope for practical research and development.

% The proposed novel competition design leads to an interesting confrontation dynamic. 
% It outlines the importance of model concealing and a model breakage timeline and can be reused in other domains and modalities to gain valuable insights. 
% Additionally, our study uncovers simple yet effective attack and defense methods that surpass existing alternatives. 
% These strategies, including a defense mechanism based on filtering suspicious events, are made available in an open repository for broader application and further research.

Banks routinely use neural networks to make decisions.
While these models offer higher accuracy, they are susceptible to adversarial attacks, a risk often overlooked in the context of event sequences, particularly sequences of financial transactions, as most works consider computer vision and NLP modalities. 

We propose a thorough approach to studying these risks: a novel type of competition that allows a realistic and detailed investigation of problems in financial transaction data.
The participants directly oppose each other, proposing attacks and defenses --- so they are examined in close-to-real-life conditions.

The paper outlines our unique competition structure with direct opposition of participants, presents results for several different top submissions, and analyzes the competition results. 
We also introduce a new open dataset featuring financial transactions with credit default labels, enhancing the scope for practical research and development.
\end{abstract}


\maketitle

\section{Introduction}
The evolution of the modern financial sector has been marked by rapid advancements in technology, enabling financial institutions to offer better services with improved efficiency. 
One of the main contributors over the last decades is machine learning~\cite{dixon2020machine}. 
Enhanced model quality built on timely and accurately collected data leads to improvement in quality and  
decision-making speed in banks~\cite{tripathi2012survey}.
However, these advancements have simultaneously opened up new channels for malicious actors to exploit, one of which is the emergence of adversarial attacks on machine learning models~\cite{szegedy2014intriguing}.
The issue becomes even more pressing in the context of financial transaction data, where the stakes have explicit monetary value, and robust defense mechanisms are needed~\cite{fursov2021adversarial}.

Financial transaction data consist of sequences of transactions produced by customers.
While close to natural language~\cite{qiu2022adversarial} and event sequences data~\cite{khorshidi2022adversarial,shchur2021neural}, this modality has notable differences. It includes, in particular, dependence on macroeconomic situation, higher required attention ranges, and higher diversity of available features~\cite{babaev2022coles}, thus implying a separate line of research on financial transaction data robustness. 

% Understanding adversarial strategies and developing countermeasures against them has been a growing area of interest but remains underexplored in the realm of financial transaction data. Research in this field has largely been confined to static environments and traditional data modalities, leaving financial transaction data relatively uncharted.

One possible way to explore robustness is to hold competitions~\cite{goodfellow2013challenges} or maintain benchmarks~\cite{dong2020benchmarking}.
Competitive evaluation has emerged as an effective way to measure and foster advancements in machine learning~\cite{goodfellow2013challenges}.
We also see notable benchmarks on adversarial robustness~\cite{dong2020benchmarking}.
However, current approaches overlook the two-side dynamics of adversarial attacks and defenses and tend to ignore the unique challenges posed by financial transaction data~\cite{croce2021robustbench}. 
Moreover, they don't consider the full matrix of pairs of attacks and defenses against each other, making the comparison incomplete.
% TODO: comparison to NLP

% Our claims:
% \begin{itemize}
%     \item We present a competition scheme for the evaluation of the robustness of machine learning models. The unique part of the competition is the tournament phase, where participants compete against each other, looking for vulnerabilities in their rivals' models and improving the reliability of their models.
%     \item Our analysis of the dynamic of the competition reveals how sustainable the models are. We select the main factors that affect the quality of attack and defense.
%     \item We found that the models of financial transaction data exhibit different behavior and need separate algorithms for their attacks and defense compared to their computer vision and NLP counterparts.
% \end{itemize}

Given this solid background, we aim to introduce an approach to advancing the development of robust models for processing financial transaction data. 
Our primary contributions are:
\begin{itemize}
    \item A competition design: we propose a competition framework to evaluate the robustness of machine learning models in two phases. The pre-tournament phase allows for detailed study of a static model environment, while the tournament phase encourages participants to actively probe and defend against vulnerabilities, simulating real-world scenarios and enhancing the reliability of the models. The framework can be reused for other data modalities. 
    All the materials of the competition can be found at \href{https://vorsineo.github.io/adv_ml_tournament/}{https://vorsineo.github.io/adv\_ml\_tournament/}.
    % The competition was held on the special platform: \href{https://ods.ai/tracks/data-fusion-2023-competitions}{https://ods.ai/tracks/data-fusion-2023-competitions}.
    %In addition, we release a new open dataset of sequences of financial transactions with the credit default as the target.
    
    \item A new open dataset: we introduced a new unique dataset on financial transactions with a credit default target, crucial for banks. This target is different from targets in other openly available datasets.
    %The dataset is available at \href{}{}.
    
    \item A dynamics analysis: we collected and tested top participants' submissions of the real financial transaction data for the introduced dataset and another open one. During the analysis, we demonstrated that financial transaction data requires specialized algorithms for attacks and defenses. In particular, a new defense based on the identification of suspicious events was proposed. Also, we examined the possible alternative random forest model as a more robust one. 
    % The proposed approaches' superiority is based on evidence collected during experiments with real financial transaction data for the introduced dataset and another open one.
    % \item A dynamics analysis: through analyzing the dynamics of the competition, we uncover insights into model sustainability and identify crucial factors influencing attack and defense quality. Such analysis enables us to gauge the models' real-world efficacy.
    
    % \item Baseline attack and defense strategies for financial transactions: we demonstrate that financial transaction data require specialized algorithms for attacks and defenses, differentiating them from their counterparts in computer vision and NLP. In particular, we propose a new defense based on the identification of suspicious events and examine the possible usage of random forests as a more robust model.
    % The proposed approaches' superiority is based on evidence collected during experiments with real financial transaction data for the introduced dataset and another open one.
\end{itemize}

The rest of this paper is structured as follows.
% \ref{sec:related}
Section~\ref{sec:related} is devoted to related work on the topic.
% \ref{sec:data}
In Section~\ref{sec:data}, we describe the presented dataset and its structure.
% \ref{sec:competition}
Section \ref{sec:competition} proposes a novel competition structure.  
% \ref{sec:experiments}
Finally, Section~\ref{sec:experiments} delves into the analysis of the competition's dynamics and the findings related to the robustness of the models, as well as the comparison of developed attacks and defenses to existing baselines.

\section{Related work}
\label{sec:related}
\begin{figure}
    \centering
    \includegraphics[width=\columnwidth]{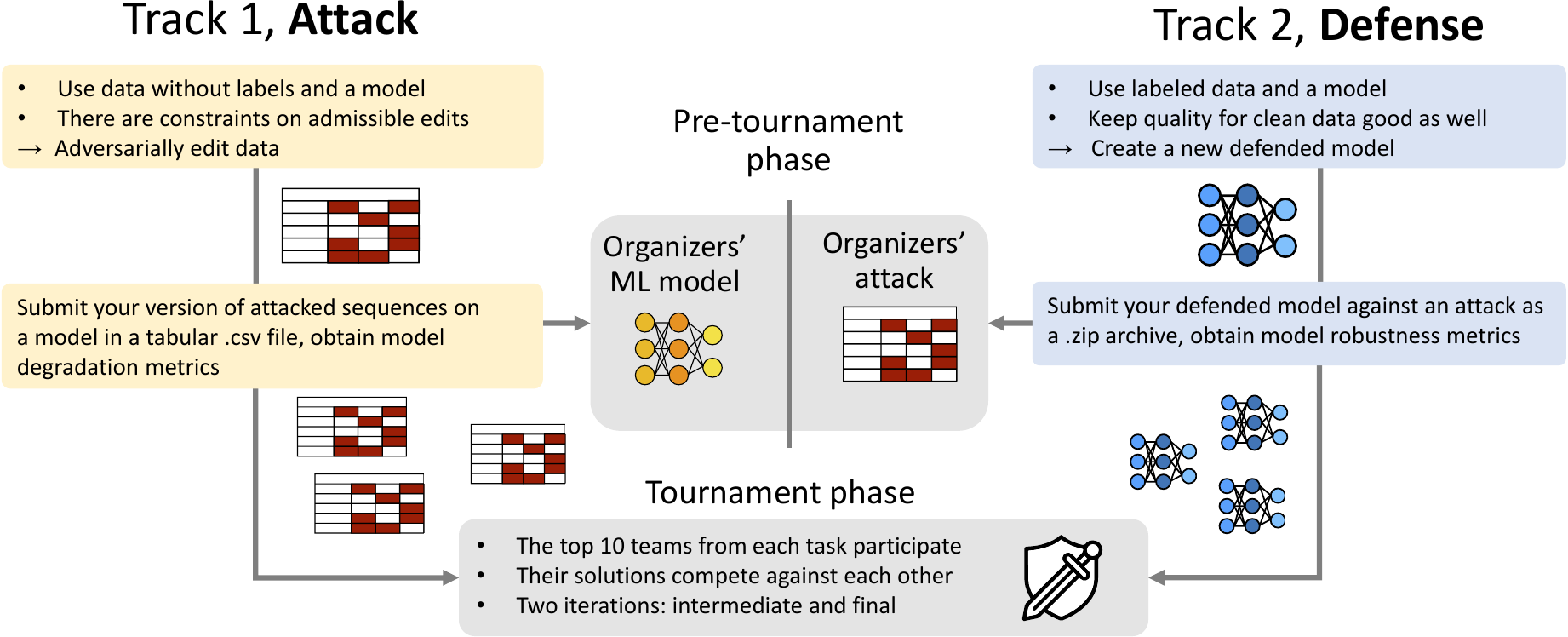}
    \caption{Competition scheme with attack and defense tracks and pre-tournament and tournament phases. Better to view in zoom}
    \label{fig:tournament_scheme}
\end{figure}

The finance sector remains a prime target for malevolent actors.
Adversarial actions lead to significant losses to banks and their customers~\cite{tripathi2012survey}.
With the broad adoption of complex machine learning models in banking, the industry should design new risk management opportunities. 
In particular, the emergence of adversarial attacks on such models poses challenges that require urgent attention.
In light of this, there has been a growing interest in studying the mechanisms behind adversarial attacks and developing defense systems, as well as in training more sophisticated models to process such data.

\paragraph*{Models of Financial Transactions Data}

Sequences of financial transaction data offer a comprehensive understanding of a client's behaviors.
Neural networks (NNs) have been widely adopted in this area~\cite{BANYMOHAMMED2024100215, BUENO2024100230, SINGH2022100094, jurgovsky_sequence_2018, AMATO2024100234, kaya2019artificial} due to their ability to process large-scale, complex sequences with high performance without intermediate steps of feature generation. 
Research indicates the utilization of convolutional (CNN)~\cite{fursov2021adversarial} and recurrent neural networks (RNN)~\cite{babaev2019rnn} for prediction tasks, including credit scoring and churn detection. 
% The connection between clients can also be explored via the adoption of graph neural networks~\cite{sukharev2020ews}, as these connections provide a rich source of information about their default behaviour~\cite{munoz2023combination}.
The presented neural networks can also serve as encoders, providing representations suitable for solving numerous problems~\cite{babaev2022coles}.
We also note a connection between event sequences often described as temporal point processes and sequences of financial transactions, as papers in this area routinely use such datasets as a part of their methods' evaluation protocol~\cite{zhuzhel2023continuous}.
Moreover, financial transactions from major bank clients are a good indicator of macroeconomic trends, making possible predictions of diverse macroeconomic indexes~\cite{begicheva2021bank}.
So, many decisions in banks, including credit scoring and overall strategy, rely on neural network-based processing of financial transaction data~\cite{ala2022deep,wang2022deep}.

\paragraph*{Adversarial Attacks and Defenses}

As such, the usage of neural network models expands in the area of financial transaction data despite known vulnerabilities, such as adversarial attacks.
It is long known that small adversarial perturbations of input to NN can lead to significant changes of output~\cite{szegedy2014intriguing}, often completely disrupting model predictive power.
A taxonomy of attacks~\cite{tabassi2019taxonomy} and recent reviews~\cite{xu2020adversarial} mention several types of attacks, such as evasion, poisoning, and reverse engineering attacks. 
There are now a number of approaches that have become baselines in this area~\cite{pgd, fsgd, autoattack, cw_attack}.
Meanwhile, the defense strategies range from adversarial training and defensive distillation to feature squeezing and ensembling. 
These defenses fortify models against adversarial interference by enhancing their ability to resist perturbations or detect and nullify an attack. Recently, various defense methods have been developed and their theoretical and practical properties have been considered~\cite{rew1, rew2, rew4, rew5}. 
% ~\cite{pgd, fsgd, autoattack, cw_attack}. 
The horizons of applications of machine-based attacks continue to expand in recent years~\cite{jueguen2023broadening}.
One direction is to explore the vulnerability of different types of models, including decision trees~\cite{vos2022robust} and logistic regression~\cite{dan2020sharp}.
Another option is to consider different application areas.
In particular, adversarial attacks have gained significant attention in the financial sector due to the critical implications of successful breaches~\cite{kumar2021evolutionary} even for tabular imbalanced data~\cite{cartella2021adversarial}. 
The paper~\cite{fursov2021adversarial} considered adversarial attacks and defenses for financial transactions data modality. 
They proposed to use a gradient-enhanced generative model to create an attack and considered discriminator-based and adversarial training-based defense strategies. 
Other approaches for attacking event sequence models are presented in~\cite{khorshidi2022adversarial,kovtun2023hiding}.
To sum up, this attacker-defender opposition continues, leading to the advancement of both robustness and attack strength. 
However, research on attacks and defenses for financial transaction data is relatively scarce.
We still don't understand to what extent the models can be broken and how harmful malicious actions can be, requiring a fast track for such research in the face of presented threats.

\paragraph*{Competitions in machine learning}

Competitions have emerged as a powerful tool to drive innovation and accelerate progress in machine learning. 
One of the pioneering works examining the importance of contests in machine learning is presented in the work~\cite{goodfellow2013challenges}.
It discusses how one should design competitions and how the results move forward research in machine learning, fostering a collaborative and competitive environment that prompts participants to devise novel methods and algorithms. 
The current state-of-the-art in machine learning competitions is presented in~\cite{carlens2023state}.
However, most competitions focus on improving the performance of the models, paying little attention to the robustness of proposed solutions.
Moreover, in present contests, participants don't compete against each other, and all interactivity at available platforms is constrained to RL-related contests.

An alternative to competitions is benchmarks that collect metrics from various papers, the most well-known being \emph{paperswithcode}~\footnote{\url{https://paperswithcode.com/}}.
Specifically in the context of adversarial attacks and defenses, the article~\cite{dong2020benchmarking} proposed adversarial robustness benchmarks for computer vision problems that were utilized in a competition~\cite{dong2023competition}. 
Another essential benchmark of adversarial defenses was presented in~\cite{croce2021robustbench}.
The authors evaluated several models against the AutoAttack approach for computer vision problems.
However, due to apparent constraints, competitions are often limited to a static environment without confrontation of participants. 

\paragraph*{Research Gap}

Despite the significant advancements in understanding adversarial attacks and defenses, a gap persists in financial transaction data. 
Available research focuses on other data modalities and types of models. 
Furthermore, although various machine learning competitions appear, they almost pass over financial transaction data, work in a static environment, and ignore the competitive nature of adversarial studies. We aim to bridge this gap by developing a competition to explore adversarial strategies in financial systems and evaluate their countermeasures to enhance their robustness.
Due to the proposed competition design, we can identify the best attacks and defenses, improving our understanding of constructing robust models based on financial transaction data. 

Another gap is the lack of data in this research area. Since bank users' transaction data contains personal information and is also used for decision-making systems in banks, the owners are not interested in making it publicly available. We mitigate this gap by releasing the new public anonymized dataset of bank transactions. This dataset also contains the credit default target, which was not considered in the previous open datasets. 

% \section{Methods}
\label{sec:methods}

% In this section, we discuss the peculiarities of financial data and the structure of the proposed competition. Also, we describe methods used in the comparison in the experiments section.

\section{Data}\label{sec:data}

\subsection{General transactional data overview}
The financial transaction data can be represented as event sequences, where each event is one transaction, and each sequence is a sequence of transactions from one user of a bank. Such data have certain differences in comparison with regular time series. 
The main difference is non-uniformity: the time passed between subsequent transaction events varies. 
Also, a description of each event is multidimensional, with each dimension being either continuous or discrete. 
For example, info on each transaction in considered datasets includes merchant category code (MCC) and amount.
These two features are among the most critical indicators of customer behavior~\cite{curry2007detecting}.
The MCC is a categorical feature that shows the type of transaction. 
The amount is a continuous feature describing how much money a user spent in the transaction. 
 
For such data, one is interested in classifying clients according to bank needs. 
For example, we can come up with a prediction of whether the user will leave this bank or whether the user will cease credit payments, experiencing credit default, or not. 
The main purpose of the proposed competition is to explore effective attacks and defenses for models trained for such classification tasks.
% Given the data description above, we can define a competition.

\subsection{Datasets used in competition}

For this competition, we present a new open dataset of bank transactions named Default, which is described below. Also, we applied a previously existing dataset named Churn for additional testing of the best solutions. The detailsared in appendix \ref{sec:res_churn}.

% \subsubsection{Default}

% The competition uses a year's worth of anonymized financial transaction sequences.
The Default dataset was published during this competition and can be found at \href{https://vorsineo.github.io/adv_ml_tournament/#subsection4}{https://vorsineo.github.io/adv\_ml\_tournament/\#subsection4}.
Each transaction's info includes the merchant category code (MCC), amount, currency, and time. 
The MCC belongs to about $1000$ categories, like ATM cash or drug store visits.
%The dataset consists of approximately x million transactions from y users.
Each customer within the dataset has at least 300 associated transactions in a sequence. 
For the sake of privacy, all the user's names were replaced by their identification numbers and all transaction amounts within the dataset have been anonymized with normalization and small noise.
% all the user's names were replaced by their identification numbers, and 
% выход на просрочку 90 дней в течение года
The sequences are complemented with a credit default target whether the customer failed to pay out a credit, which is the new type of objective for open datasets in this area.
For such a binary classification problem, the share of the positive label rate of defaults in data is $0.04$.
For competition purposes, we split all sequences into folds, so at each phase, participants work with a newly revealed fold.
The size of each fold ranges from $4000$ to $7000$ sequences. The complete data separation pipeline for the various stages of the competition can be found in the appendix \ref{sec:data_separation}.
% unused, bank model training, public attack, private attack, privacy protection, public protection, additional training by participants, data for tournament 1, and data for tournament 2.
% неиспользованные обучение модели банка 
% атака публичная атака приватная
% защита приватная защита публичная
% дообучение участниками данные для турнира 1 данные для турнира 2
% Furthermore, we resampled the initial dataset for more robust evaluation of metrics and avoiding significant dissimilarities between final test data and a part of it used for training. 
% TODO how we resample?

% ADD that we are closer to the real life scenario with imbalanced classes

% \paragraph{Cohorts available for participants}
% % Олег кажется что вся тема с 9 кохортами нихера не понятна, мб в терминах train test public private? Alex+++
% % Сказать что в атаке не было лейблов
% The cohort for adversarial attack design consists of 4200 customers without labels. 
% The cohort for the refinement and defense design has 7080 customers with labels. 
% Furthermore, to ensure consistency, the same time zone has been maintained across all transaction data points. 

\section{Competition flow}\label{sec:competition}
\subsection{Problem statement}
This work presents a pipeline for testing the model's security. 
The pipeline should be adaptable to different data modalities and be as close as possible to real-life scenarios.
We state that a prospective approach here is a tournament: it is close to real life and, by design, leads to rival competitors validating attacks and defenses. 
The challenge here is to provide a steady flow of attack-defense comparisons that keep the involvement high and allow for multiple attempts.

For this aim, we suggest the competition scheme, which includes the direct opposition between attacks and defences.
The proposed competition structure is depicted in Figure~\ref{fig:tournament_scheme}. 
It encompasses two distinct phases across two tracks: \emph{a preliminary phase} and \emph{a subsequent tournament phase}. 
The preliminary phase features separate \emph{tracks for attack and defense} competitions and is pretty common for adversarial benchmarks. 
The tournament phase proceeds in an innovative head-to-head format. 
The organization of this section reflects this dual-phase competition structure.

\subsection{Attack track}

% https://ods.ai/tracks/data-fusion-2023-competitions/competitions/data-fusion2023-attack
In the attack track, the participants develop an attack that significantly changes the output of a given model $f(x)$ after a minor change of input $x$, where the input is a sequence of financial transactions and the output is the probability of positive label in default classification task. 
The attacked model is a gated recurrent unit (GRU)-based model~\cite{chung2014empirical,babaev2019rnn}, the details on it are in the Appendix~\ref{sec:base_model}.
The organizer gives participants a set of sequences $X = \{x_i\}_{i = 1}^n$.
Participants provide their version of attacked sequences $X' = \{x'_i\}_{i = 1}^n$ given a constraint of the number of changed events $c(x_i, x'_i) \leq c$.
The goal is to decrease the model's ROC AUC for $X'$.

During the pre-tournament phase, an attacker has full access to the initial model, making this phase a white-box scenario.
During the tournament phase, the model is unknown to participants, as they attack models designed by others, making this phase a black-box scenario.

% Below, we give details on the evaluation metrics and the restrictions for attacks.

\subsubsection{Evaluation}

Automatic evaluation starts with a sanity check to see if modified inputs satisfy change constraints.
Then, the score is calculated as the difference in ROC AUC value between the model's predictions for the initial sample and the adversarially altered sample.
To make conditions fairer, during the competition, participants observe the results only on one-half of the test set (a public part), while the final ranking is identified via another half of the test set (a private part).

\subsubsection{Restriction for participants' attacks}

We imposed constraints to preserve the authenticity of a sequence, as banks often have models to detect and ignore fake transactions. 
Specifically, an attack can change up to ten transactions per client.
The permissible amounts for these transactions are within the minimum and maximum for the considered MCC. 
To avoid issues with boundary ambiguities, we reduced this interval with a coefficient of $0.95$. 

\subsubsection{Baseline attack}
% – Получаем разметку из модели
% – Берём самого яркого представителя из 0 и 1 классов
% – Берём у них последние транзакции,
% — для тех, у кого класс 0, подставляем транзакции класса 1 и наоборот
% — учитываем знак транзакции исходного пользователя TODO not clear
% Атака лишь немного влияла на модель (примерно 0.69 ROC AUC).
As a baseline, we adopt a simple attack.
We identify two representative customers with the highest attacked model scores for both classes.
Then, the last ten transactions from the representative customers are added to sequences for customers from opposite classes to alter the prediction for them. 
Despite this change, the impact on the model was relatively modest, with the model maintaining a high ROC AUC score of $\approx 0.69$. 
This~suggests~that~the~model~gives~fairly~accurate~predictions~in~the~presence~of~the~baseline~attack.

\subsection{Defense track}

Similar to the attack task, participants have access to a GRU binary classification model that predicts client default. 
The competition also provides access to a small labeled sample of client data.
The exact format in which this model will be attacked is an alternation of a small number of transactions in a sequence fed into the model.
The task is to construct a robust model for the same classification problem, making it resilient to such vulnerabilities. 
% As it is hard to prevent data leaks completely, it is entirely within a data scientist's capabilities to protect the model from adversities.

\subsubsection{Evaluation}
% n.b. we defend from a relatively weak attack at this phase
We calculate the ROC AUC values for model predictions for two samples, clean data and attacked data, as our aim is to avoid significant quality degradation typical for robust models~\cite{lecuyer2019certified}.
The harmonic mean of these ROC AUCs is the final defended model's quality metric.
% We compare this metric with the metric obtained via a baseline model trained without any defense capabilities in mind.

\subsubsection{Baseline for the defense track}
% Бейзлайн защиты для участников
% - Семплируем 90% транзакций каждого пользователя
% - Делаем предикт
% - Делаем так 9 раз
% - Усредняем результат для каждого пользователя
% Защита улучшала совокупное качество и бейзлайн атаки не влиял на эту модель.

For a baseline defense, we create a lightweight ensemble.
We randomly sample 90\% of transactions from an initial sequence.
We repeat this procedure $9$ times and average obtained predictions.
For this defense design, the target metric improves, reducing the effect of the baseline attack, which has negligible adversarial properties for this defense.

\subsection{Tournament phase}

To enhance the development of advanced approaches, we introduce joint attack-defense tournament phases for each track. This includes both an attack track tournament and a defense track tournament.

We select the top 10 participants from the attack track and the top 10 from the defense track, resulting in 20 solutions for each track since each participant presents two solutions.

For the attack track, participants provide a modified list of transaction sequences. For the defense track, participants present an updated model packaged in a Docker container. We evaluate each pair of attack and defense solutions to create a score matrix, as defined in the pre-tournament phase.

Finally, we rank the attack and defense solutions based on their average scores

In this phase, we conduct two iterations of the attack-defense game.
For the second one, the observed improvement is negligible, so we keep the number of iterations to two.
However,~we~can~conduct~multiple~rounds~to~see~gradual~improvements~in~attacks~and~defenses~in~black-~or~grey-box~scenarios.

\section{Results}
\label{sec:experiments}
\subsection{Competition results}

% We split our results into two parts.
% The first part is devoted to evidence obtained from competition results on the dynamic of it and presented solutions.
% The second part is about wider experiments aimed at validating conclusions from the competition in a more traditional setting.

\begin{figure*}[t!]
    \centering
    \begin{subfigure}[t]{0.248\textwidth}
        \centering
        \includegraphics[width=\textwidth]{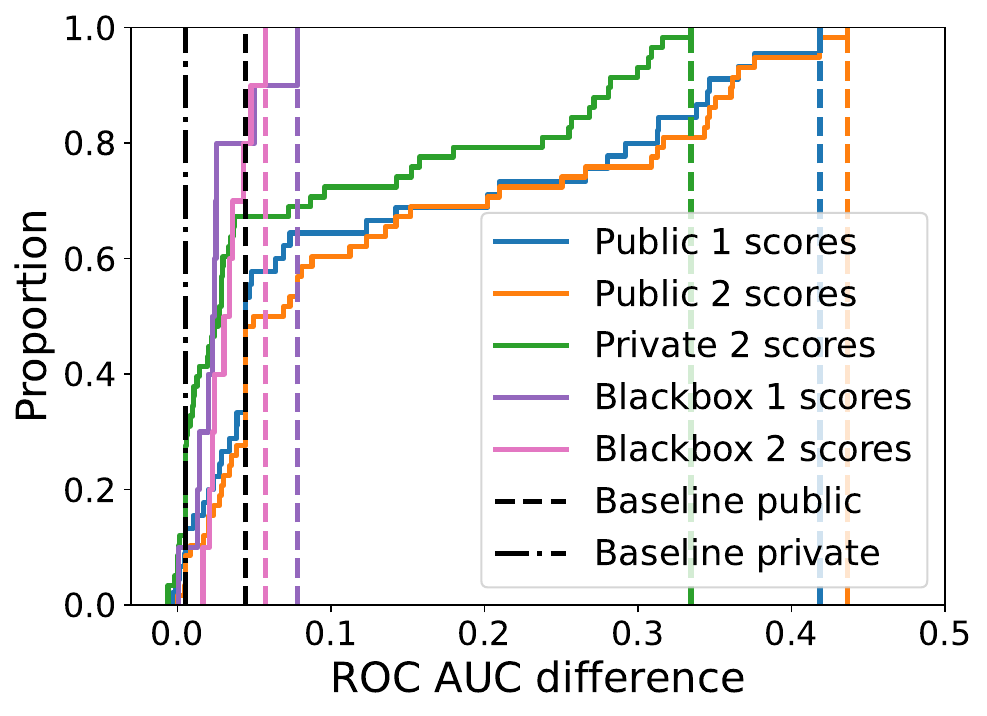}
        \caption{Attack scores}
        \label{fig:attack_scores}
    \end{subfigure}%
    ~ 
    \begin{subfigure}[t]{0.248\textwidth}
        \centering
        \includegraphics[width=\textwidth]{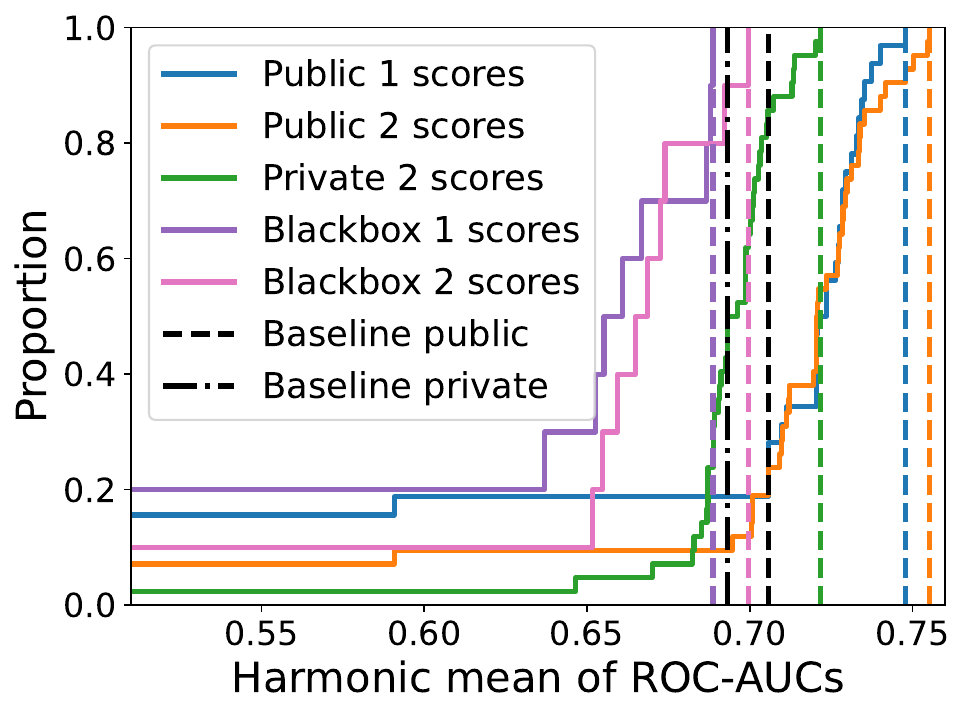}
        \caption{Defence scores}
        \label{fig:defense_scores}
    \end{subfigure}
    ~
     \begin{subfigure}[t]{0.225\textwidth}
        \centering
        \includegraphics[width=\textwidth]{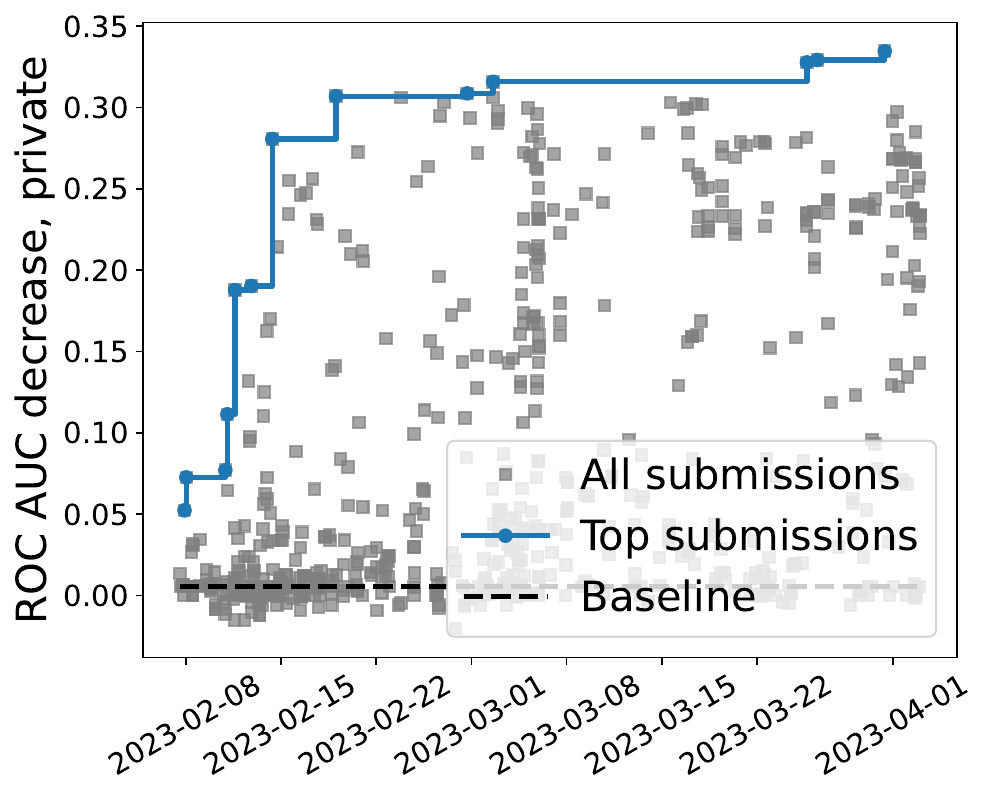}
        \caption{Attack dynamic}
        \label{fig:attack_dynamic}
    \end{subfigure}%
    ~ 
    \begin{subfigure}[t]{0.235\textwidth}
        \centering
        \includegraphics[width=\textwidth]{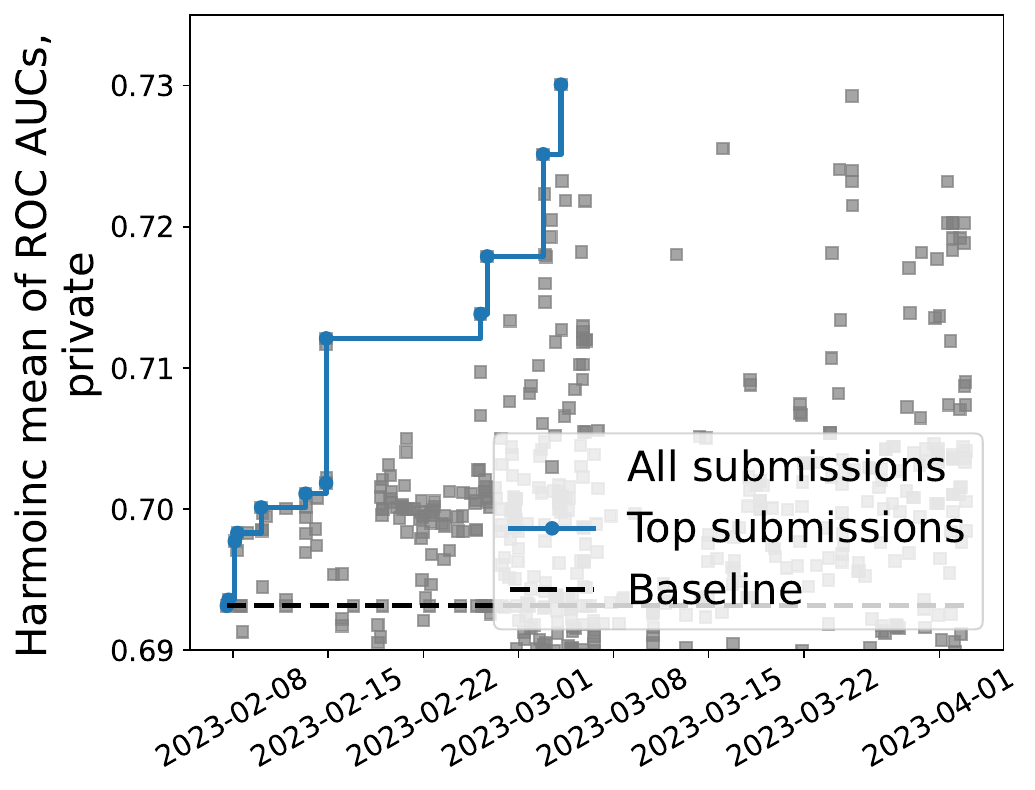}
        \caption{Defense dynamic}
        \label{fig:defense_dynamic}
    \end{subfigure}
    \caption{
    On the left pictures \ref{fig:attack_scores} and \ref{fig:defense_scores} solid lines are empirical cumulative density functions for different stages of the attack \ref{fig:attack_scores} and defense \ref{fig:defense_scores} tracks of the competition. Dashed lines are the top score for each phase (1 or 2) and baseline scores. Higher ROC AUC differences are better for an attack. Bigger Harmonic means of ROC AUCs are better for defence. On the right pictures \ref{fig:attack_dynamic} and \ref{fig:defense_dynamic} dynamic of the top private score for attack \ref{fig:attack_dynamic} and defense \ref{fig:defense_dynamic} competition is presented. These scores were hidden from participants until the end of the competition. With a blue curve, we highlight the top achieved score at a given moment in time, and each point corresponds to a score for a single submission. Better to view in zoom.}
\end{figure*}

\subsubsection{Attack quality}

% Paragraph, number of participants, their level etc.
$58$ participants took part in the competition's attack phase.
Among them were teams with strong experience in machine learning competitions.
The activity of teams resulted in $649$ submitted solutions.
During the competition among attacks, we had two control points in time to observe the progress.

In Figure~\ref{fig:attack_scores}, we present the empirical cumulative density function (CDF) for all collected scores during different phases. 
Firstly, the plot contains the top scores for each participant for a white-box scenario, with two consecutive public scores \emph{Public 1, Public 2} and the final score \emph{Private 2} obtained using a hold-out sample. 
Also, we provide results for two iterations for a black-box part of the competition \emph{Blackbox~1, Blackbox~2}.

% \begin{figure*}[h!]
%     \centering
%     \begin{subfigure}[t]{0.45\textwidth}
%         \centering
%         \includegraphics[width=0.87\textwidth]{figures/attack_dynamic.pdf}
%         \caption{Attack dynamic}
%         \label{fig:attack_dynamic}
%     \end{subfigure}%
%     ~ 
%     \begin{subfigure}[t]{0.45\textwidth}
%         \centering
%         \includegraphics[width=0.87\textwidth]{figures/defense_dynamic.pdf}
%         \caption{Defense dynamic}
%         \label{fig:defense_dynamic}
%     \end{subfigure}
%     \caption{Dynamic of the top private score for attack \ref{fig:attack_dynamic} and defense \ref{fig:defense_dynamic} competition. These scores were hidden from participants until the end of the competition. With a blue curve, we highlight the top achieved score at a given moment in time, and each point corresponds to a score for a single submission.}
% \end{figure*}

Finally, we expand our analysis with the dynamic of the top score in Figure~\ref{fig:attack_dynamic}.
For a white-box scenario, the ROC AUC started at around $0.69$. 
Following a targeted attack, this score decreased to $\approx 0.25$, making the model inoperative. 
Contrastingly, in the black box scenario, even the leading attacks only slightly impact the performance.
% the resulting ROC AUC is higher than 0.5 
This suggests that for a truly effective attack, white-box access to the model is necessary.
Additionally, the dynamics of the score change indicate that after gaining access to a model, it might only take about two weeks to compromise it completely. 
Therefore, it's crucial for model owners to act swiftly after a model leakage.
% todo: what is going on with the scores during ?

\subsubsection{Defense quality}

For a less traditional defense track, the number of participants was $42$, while the strongest participants submitting to both tracks.
Here, we also present the results for different stages: two stages for a given attack \emph{Public 1, Public 2, Private 2} and two tournament stages \emph{Blackbox 1, Blackbox 2} for unknown attacks authored by other participants. 
We also present baseline scores for the tournament phase \emph{Tournament public} and \emph{Tournament private}.

The empirical CDF for the defense track is in Figure~\ref{fig:defense_scores}.
We expand our analysis with the dynamic of the top score in Figure~\ref{fig:defense_dynamic}.

Clearly, one can improve over the baselines if the model hasn't been protected before.
Part of the improvement comes from the improvement of the model quality for clean data, as we use the harmonic mean of the quality for clean and attacked data for evaluation.
While weaker attacks lead to minor effects, with the highest score of $0.72$ being higher than the quality of the initial model, the defenses significantly degrade when put against stronger attacks.
For tournament phases, the protected models' scores are close to baselines.
We also note a slight improvement when comparing the first and the second checkpoints.
The model would be significantly better defended after two weeks of effort, making it an estimation of how much time we need to break the model.  
Finally, during the last month of the competition, there was no improvement, suggesting that the models reached protection from the baseline attack.

\begin{figure}[b!]
    \centering
    \includegraphics[width=0.6\linewidth]{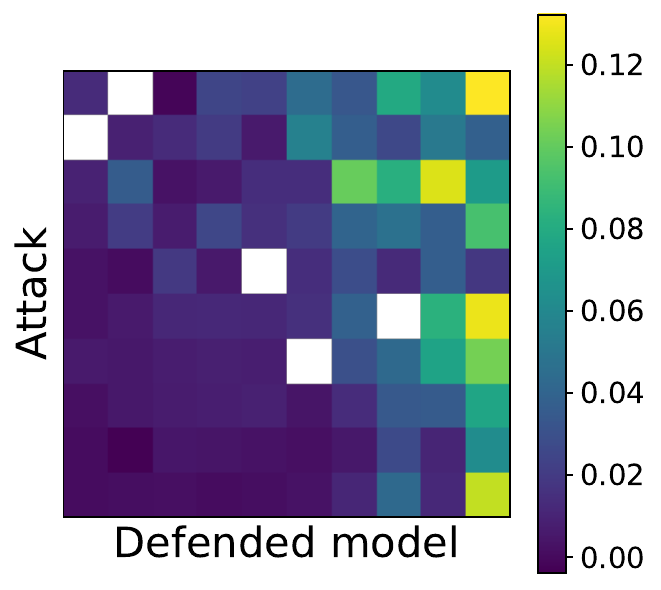}
    \caption{ROC AUC decreases for pairs of attack and defense from the competition stage. We removed unfair scores for pairs when the attack and defense model authors coincided and put instead white squares.}
    \label{fig:comp_matrix}
\end{figure}

\subsubsection{Defence versus attack quality}

During the black box tournament stage, participants submit their defended models and attacked transactions sequences, standing against each other.
There were two tournament stages, but the results for them are similar, so we show details only for the second~one.

The ROC AUC decreases for each pair of attack and defense are in Figure~\ref{fig:comp_matrix}.
We also provide scores for all attacks and defenses sorted by their median values in Appendix \ref{sec:roc_auc_decrease}. 
The solutions significantly differ in quality. 
Despite the pure black-box nature of this phase, we still see defenses and attacks with almost zero decrease in ROC AUC and about a $0.05$ decrease of ROC AUC on average, suggesting that even in this regime, one can defend and attack.
Moreover, the attack transfers, as ROC AUC degradations of defended models correlate.

\begin{table}[b]
    \centering
      \begin{tabular}{lcc}
      \hline 
        Model & Public & Private \\
        \hline
        NN base & 0.7035 & 0.6876 \\
        NN mix & 0.7134 & 0.6960 \\
        Boosting base & 0.7415 & \textbf{0.7279} \\
        Boosting mix, 2 & 0.7403 & 0.7255 \\
        Boosting mix, 5 & \underline{0.7519} & \underline{0.7221} \\
        Boosting mix filter & \textbf{0.7529} & 0.7197 \\ 
        Boosting clean & 0.6883 & 0.6309 \\
        \hline
        \end{tabular}
        
        \caption{ ROC AUC values for considered defended models for clean data}
        \label{table:model_performance}
\end{table}

% How transfer works (can we identify pairs of attack and defence produced by the same team?)
\begin{table*}[t]
    \centering
    \begin{tabular}{lcccccccccc}
    \hline
    
    Model & No  &  Random & NN  & Boost.  & Boost.  & NN base  &  NN mix  & NN base + &   NN base + & Mean \\
    & attack & & base & base &  mix alt &  gradient &  gradient &  Boost. base &  Boost. mix alt &  \\
    \hline
    NN base & 0.7035 & 0.7007 & \underline{0.3958} & 0.6924 & 0.6884 & \underline{0.3343} & 0.6607 & 0.6720 & \underline{0.4399} & \underline{0.5548} \\
    NN mix & 0.7134 & 0.7192 & 0.7048 & 0.7105 & 0.7127 & 0.7134 & 0.6976 & 0.7108 & 0.7076 & 0.7082 \\
    Boosting base & 0.7415 & 0.7269 & 0.7188 & \underline{0.3783} & 0.6038 & 0.7405 & 0.7017 & \underline{0.3780} & 0.6670 & \underline{0.5983} \\
    Boosting mix, 2 & 0.7432 & 0.7383 & 0.7408 & \underline{0.4250} & 0.7010 & 0.7458 & 0.7333 & 0.7251 & \underline{0.4275} & 0.6426 \\
    Boosting mix, 5 & 0.7519 & 0.7457 & 0.7473 & \underline{0.5347} & 0.7247 & 0.7544 & 0.7426 & \underline{0.5374} & 0.7361 & 0.6825 \\
     Boosting mix filter & 0.7529 & 0.7461 & 0.7556 & 0.7157 & 0.7509 & 0.7519 & 0.7425 & 0.7148 & 0.7528 & 0.7406 \\
    \hline
    Mean & 0.7344 & 0.7295 &	0.6772 & \underline{0.5761} &	0.6969	& 0.6734 & 0.7131 & 0.6230 & 0.6218 & 0.6545 \\
    \hline
    \end{tabular}
        \caption{ROC AUC values for various attacks and models for the initial \emph{Default} dataset and mean values over rows and columns. Rows correspond to different defended models, and columns correspond to different attacks. Here, we underline $\leq 0.6$ ROC AUC scores corresponding to successful attacks.}
        \label{table:perf_matrix}
\end{table*}
\subsection{Proposed attacks and defenses}

% During the competition $58$ participants took part in the attack phase. Among them were teams with strong experience in machine learning competitions.
% The activity of teams resulted in $649$ submitted solutions.
% For a less traditional defense track, the number of participants was $42$, while the strongest participants submitted to both tracks.
As a result of the competition, we obtained methods for attacks and defense that supersede existing ones.
% Their description is provided below.
% Here, we also discuss baselines that show strong results. 
The main trend among the submissions on the defence track was the usage of the different types of random forest models, including boosting that uses aggregated features. As for the attack track, the participants mainly used various brute force attacks for different types of models and also tried out several gradient-based attacks.
The description of the best solutions is provided below.
Here, we also discuss baselines that show strong results.

\subsubsection{Defended models}

The study considers six models with different defensive properties and raw data quality produced by leading teams: two neural networks and four variants of gradient boosting models.

The basic GRU-based neural network is \textbf{NN base}. 
A stronger baseline \textbf{NN mix} is the one described above as the baseline for the defense track.
% Then, we use a more robust version of this model with augmentations: we run NN $10$ times, taking $0.9$ share of transactions uniformly random without return and performing random permutation of transactions. 
% The average of the outputs is a \emph{NN mix} baseline.

In addition to neural networks, we consider four variants of gradient-boosting ensembles of decision trees.
We start with a single gradient boosting \textbf{Boosting base} with $400$ features in total aggregated from sequences of transactions.
The aggregates are mostly done via mean and sum over a single MCC.
We use CatBoost~\cite{prokhorenkova2018catboost} implementation with training via distillation from an \emph{NN mix} output, as the amount of available data in the competition is limited.
\textbf{Boosting mix, 2} adds another gradient boosting model, constructed via LightGBM with the same input features~\cite{ke2017lightgbm}.
The \textbf{Boosting mix, 5} is a weighted average of two boosting models produced via \emph{Boosting mix, 2} and three boosting models that were constructed via CatBoost without the usage of features that can be changed during an adversarial attack. 
The weights of models are $0.5$, $0.5$, $1$, $1$, and $1$ correspondingly.
Finally, for \textbf{Boosting mix filter}, we train an additional Filter classifier that identifies transactions that are likely to be changed via an adversarial attack and filters out such transactions, keeping only reliable ones.
The Filter classifier is another gradient boosting that takes a single financial transaction as an input that was trained using a subset of different attacks. 
\emph{Boosting mix fitter} and \emph{Boosting mix 5} are the approaches used by the winning team of the defense track.

% Лучший ансамбль бустингов + очистка (1 место в сореве)	добавили новый бустинг, обучен на транзакциях, не на последовательностях, а на индивидуальных транзакциях, выкидывает те, которые атакованы, порог какой-то по выкидыванию, обучаем на разных атаках
% Лучший ансамбль бустингов (2 место в сореве)	5 бустингов, 0.5 кэтбуст + 0.5 лайтгбм + 3 * 1 кэтбустинга на фичах без MCC и стоимостей (ег количество транзакций по часам)
% Лучший соло бустинг	всего 400 фичей, из них 300+ фичей - суммарные или средние стоимости по популярным МСС, дистилляция нейронки с перемешиванием
% Нейронка, 10 перемешек (выкл дропаут в голове)	бейзлайн защита, берем 0.9 транзакций случайно без возвращения, перемешиваем их, так делаем 10, потом 10 раз делаем инференс
% Нейронка (выкл дропаут в голове)	ванильна модель

\subsubsection{Attacks}

As the number of changes is limited, our attack is close to a greedy brute force approach.
At each step, we generate a preselected large number of possible substitutions of transactions and select the ones that most significantly decrease the model's score.
The number of candidates at each iteration is $1000$.
We repeat this procedure for $k$ steps, where $k$ is the number of allowed substitutions.
If not mentioned otherwise, $k = 10$.
The attack differs in the source of scores, being one of the defended models introduced above \textbf{NN base}, \textbf{NN mix}, or \textbf{Boosting base}.
Additional \textbf{Boosting mix alt} is an attack on an ensemble of gradient boosting models used by the winner of the attack track.
It combines a variety of different boosting models to generalize better to different black-box defenses.
%
% \begin{figure}[b!]
%     \centering
%     \includegraphics[width=0.5\textwidth]{figures/attack_scores.pdf}
%     \caption{Solid lines are empirical cumulative density functions of scores for different stages of the attack track. 
%     Dashed lines are the top score for each phase and baseline scores. Higher ROC AUC differences are better}
%     \label{fig:attack_scores}
% \end{figure}
%
We also consider combinations of models to try to produce stronger and varied attacks
\textbf{NN base + Boosting base} and \textbf{NN base + Boosting mix}.
For them, the average score from the two models guides the attack. 

We note that while there exists a large body of work dedicated to adversarial attacks on random forests~\cite{chen2019robust,calzavara2019adversarial}, they are not applicable in our case, as we use aggregates of transactions that can't be straightforwardly modified to take into account this peculiarity.

To enrich the space of attacks, we consider two gradient-based attacks \textbf{NN base grad}, \textbf{NN mix grad} with a similar number of changes. 
They are a variant of FGSM introduced in~\cite{fursov2021adversarial} that applies gradient in the embedding space to generate the next substitution with subsequent replacement via the closest by $l_2$ norm token.

% naive 	атаки исходной нейронки, случайно заменяя и выбирая замену, которая максимально роняет скор после замены			
% naive adv	помечены атаки, в которых использовался мой генератор замен с турнира, в котором	более реалистичный генератор замен	фокус для рнн	убрать самые сильные МСС коды
% multi cb	атака под ансамль бустингов (примерно как Лучший ансамбль бустингов)	но он отличается от другого		
% cb brute	берем лучший соло бустинг и атаку под него	 лучший соло бустинг		
% атака градиентая	прочитать сообщение			
% naive + multi	прочитать сообщение	какой-то процент меняем первой атакой, какой-то второй атакой		
% градиентная атака	первая нейронка			
% градиентная атака 2	10 перемешек нейронка

\subsection{Model performance for clean data}

Table~\ref{table:model_performance} presents ROC AUC scores for introduced approaches for private and public parts of the Default dataset before the attack. 
For the sake of comparison, we also present scores for \textbf{Boosting clean} model that uses only features not susceptible to the attack.
Firstly, the sequential nature of data can be safely ignored, as a permutation of input and aggregation after ensembling leads to performance improvement in \emph{NN mix}.
Secondly, even if distilled from a neural network model, gradient boosting shows improved performance.
The last two boosting models \emph{Boosting mix, 5} and \emph{Boosting mix filter} should stand out only in the attack scenario, as they are designed with improved robustness in mind.
We also observe that the overall ranking of models for public and private parts of the dataset is close, signifying the relative stability of the used evaluation.

\subsection{Attack and defense performance}

Table~\ref{table:perf_matrix} presents ROC AUC scores for pairs of considered attacks and defenses for the \emph{Default} dataset released with the competition.
Two additional columns here are for the performance of models with no attack \emph{No attack} and with $10$ random changes \emph{Random}.

The competition performance highly correlates with a deeper investigation conducted here, suggesting that a competition format is a viable way to develop and validate new approaches for adversarial robustness performance.
The conclusion replicates for both datasets.
In most cases, attacks harm the model's predictive power, slightly outperforming random changes.
For a white-box scenario, some attacks significantly alter the performance, leading to $<0.5$ ROC AUC.
Last but not least, gradient boosting models demonstrate better attack robustness, especially when combined with filtering and ensembling. 
We note that \emph{Boosting mix filter} provides superior performance for both datasets. 
Moreover, the introduced defense doesn't lead to model degradation for clean data cf. other methods we tried~\cite{madry2018towards,wong2019fast,liu2020adversarial}.

\section{Acknowledgments}
The work by Alexey Zaytsev was supported by a grant for
research centers in the field of artificial intelligence, provided
by the Analytical Center for the Government of the Russian
Federation in accordance with the subsidy agreement (agree-
ment identifier 000000D730321P5Q0002) and the agreement
with the Ivannikov Institute for System Programming of the
Russian Academy of Sciences dated November 2, 2021 No.
70-2021-00142.

\section{Conclusions}
\label{sec:conclusion}
While the robustness of models in areas such as Computer Vision and NLP has received extensive research attention, event sequence data, widely applied in industries such as banking, remains relatively underexplored. 
We consider this data modality and reveal a brand new dataset of bank transaction.

Furthermore, our work introduces a novel competition scheme that simulates real-world adversarial dynamics and takes into account existing constraints for attacks like those enforced by anti-fraud systems. 
This approach has revealed novel attack and defense strategies, including the first reported use of a strong neural network distillation to gradient boosting. 

Notably, our competition results highlight the vulnerability of financial models to adversarial attacks, even in a black-box context with limited transaction alterations (only $3\%$ of changes).
The most effective defense is model concealment, while other options like filtering input sequences for suspicious transactions and using more robust gradient boosting models are worth attention.  
Furthermore, we discovered that flexible competition formats yield significant insights into adversarial tactics in industrial scenarios. Our findings include the analysis of the dynamics of the breakdown and fortification of models, never explored before.
So, this work not only advances understanding in the field but also provides actionable strategies for enhancing the robustness of models handling sequential financial data.

\bibliographystyle{ieeetr}
\bibliography{references}

\begin{thebibliography}{10}

\bibitem{dixon2020machine}
M.~F. Dixon, I.~Halperin, and P.~Bilokon, {\em Machine learning in finance}, vol.~1170.
\newblock Springer, 2020.

\bibitem{tripathi2012survey}
K.~K. Tripathi and M.~A. Pavaskar, ``Survey on credit card fraud detection methods,'' {\em International Journal of Emerging Technology and Advanced Engineering}, vol.~2, no.~11, pp.~721--726, 2012.

\bibitem{szegedy2014intriguing}
C.~Szegedy {\em et~al.}, ``Intriguing properties of neural networks,'' in {\em ICLR}, 2014.

\bibitem{fursov2021adversarial}
I.~Fursov {\em et~al.}, ``Adversarial attacks on deep models for financial transaction records,'' in {\em ACM SIGKDD}, 2021.

\bibitem{qiu2022adversarial}
S.~Qiu, Q.~Liu, S.~Zhou, and W.~Huang, ``Adversarial attack and defense technologies in natural language processing: A survey,'' {\em Neurocomputing}, vol.~492, pp.~278--307, 2022.

\bibitem{khorshidi2022adversarial}
S.~Khorshidi, B.~Wang, and G.~Mohler, ``Adversarial attacks on deep temporal point process,'' in {\em IEEE ICMLA}, pp.~1--8, IEEE, 2022.

\bibitem{shchur2021neural}
O.~Shchur, A.~C. T{\"u}rkmen, T.~Januschowski, and S.~G{\"u}nnemann, ``Neural temporal point processes: A review,'' {\em arXiv:2104.03528}, 2021.

\bibitem{babaev2022coles}
D.~Babaev {\em et~al.}, ``Co{LES}: contrastive learning for event sequences with self-supervision,'' in {\em SIGMOD}, pp.~1190--1199, 2022.

\bibitem{goodfellow2013challenges}
I.~J. Goodfellow {\em et~al.}, ``Challenges in representation learning: A report on three machine learning contests,'' in {\em Neural Information Processing: ICONIP}, pp.~117--124, Springer, 2013.

\bibitem{dong2020benchmarking}
Y.~Dong {\em et~al.}, ``Benchmarking adversarial robustness on image classification,'' in {\em CVPR}, pp.~321--331, 2020.

\bibitem{croce2021robustbench}
F.~Croce {\em et~al.}, ``Robustbench: a standardized adversarial robustness benchmark,'' in {\em NeurIPS}, 2021.

\bibitem{BANYMOHAMMED2024100215}
A.~{Bany Mohammed}, M.~Al-Okaily, D.~Qasim, and M.~{Khalaf Al-Majali}, ``Towards an understanding of business intelligence and analytics usage: Evidence from the banking industry,'' {\em International Journal of Information Management Data Insights}, vol.~4, no.~1, p.~100215, 2024.

\bibitem{BUENO2024100230}
L.~A. Bueno, T.~F. Sigahi, I.~S. Rampasso, W.~{Leal Filho}, and R.~Anholon, ``Impacts of digitization on operational efficiency in the banking sector: Thematic analysis and research agenda proposal,'' {\em International Journal of Information Management Data Insights}, vol.~4, no.~1, p.~100230, 2024.

\bibitem{SINGH2022100094}
V.~Singh {\em et~al.}, ``How are reinforcement learning and deep learning algorithms used for big data based decision making in financial industries–a review and research agenda,'' {\em International Journal of Information Management Data Insights}, vol.~2, no.~2, p.~100094, 2022.

\bibitem{jurgovsky_sequence_2018}
J.~Jurgovsky {\em et~al.}, ``Sequence classification for credit-card fraud detection,'' {\em Expert Systems with Applications}, vol.~100, pp.~234--245, 2018.

\bibitem{AMATO2024100234}
A.~Amato, J.~R. Osterrieder, and M.~R. Machado, ``How can artificial intelligence help customer intelligence for credit portfolio management? a systematic literature review,'' {\em International Journal of Information Management Data Insights}, vol.~4, no.~2, p.~100234, 2024.

\bibitem{kaya2019artificial}
O.~Kaya, J.~Schildbach, D.~B. AG, and S.~Schneider, ``Artificial intelligence in banking,'' {\em Artificial intelligence}, 2019.

\bibitem{babaev2019rnn}
D.~Babaev, M.~Savchenko, A.~Tuzhilin, and D.~Umerenkov, ``{ET-RNN}: Applying deep learning to credit loan applications,'' in {\em ACM SIGKDD}, 2019.

\bibitem{zhuzhel2023continuous}
V.~Zhuzhel {\em et~al.}, ``Continuous-time convolutions model of event sequences,'' {\em arXiv:2302.06247}, 2023.

\bibitem{begicheva2021bank}
M.~Begicheva and A.~Zaytsev, ``Bank transactions embeddings help to uncover current macroeconomics,'' in {\em IEEE ICMLA}, pp.~1742--1748, IEEE, 2021.

\bibitem{ala2022deep}
M.~Ala’raj, M.~F. Abbod, M.~Majdalawieh, and L.~Jum’a, ``A deep learning model for behavioural credit scoring in banks,'' {\em Neural Computing and Applications}, pp.~1--28, 2022.

\bibitem{wang2022deep}
C.~Wang and Z.~Xiao, ``A deep learning approach for credit scoring using feature embedded transformer,'' {\em Applied Sciences}, vol.~12, no.~21, p.~10995, 2022.

\bibitem{tabassi2019taxonomy}
E.~Tabassi, K.~J. Burns, M.~Hadjimichael, A.~D. Molina-Markham, and J.~T. Sexton, ``A taxonomy and terminology of adversarial machine learning,'' {\em NIST IR}, vol.~2019, pp.~1--29, 2019.

\bibitem{xu2020adversarial}
H.~Xu {\em et~al.}, ``Adversarial attacks and defenses in images, graphs and text: A review,'' {\em International Journal of Automation and Computing}, vol.~17, pp.~151--178, 2020.

\bibitem{pgd}
A.~Madry, A.~Makelov, L.~Schmidt, D.~Tsipras, and A.~Vladu, ``Towards deep learning models resistant to adversarial attacks,'' in {\em ICLR}, 2018.

\bibitem{fsgd}
I.~J. Goodfellow, J.~Shlens, and C.~Szegedy, ``Explaining and harnessing adversarial examples,'' in {\em ICLR}, 2015.

\bibitem{autoattack}
F.~Croce and M.~Hein, ``Reliable evaluation of adversarial robustness with an ensemble of diverse parameter-free attacks,'' in {\em ICML}, 2020.

\bibitem{cw_attack}
N.~Carlini and D.~Wagner, ``Towards evaluating the robustness of neural networks,'' in {\em IEEE Symposium on Security and Privacy (SP)}, 2017.

\bibitem{rew1}
M.~Terzi, A.~Achille, M.~Maggipinto, and G.~A. Susto, ``Adversarial training reduces information and improves transferability,'' in {\em AAAI}, 2021.

\bibitem{rew2}
D.~Zhou {\em et~al.}, ``Improving adversarial robustness via mutual information estimation,'' in {\em ICML}, 2023.

\bibitem{rew4}
D.~Zhou {\em et~al.}, ``Removing adversarial noise in class activation feature space,'' in {\em ICCV}, 2021.

\bibitem{rew5}
M.~Lee and D.~Kim, ``Robust evaluation of diffusion-based adversarial purification,'' in {\em CVPR}, 2023.

\bibitem{jueguen2023broadening}
J.~V. Jueguen, {\em Broadening the Horizon of Adversarial Attacks in Deep Learning}.
\newblock PhD thesis, Universidad del Pa{\'\i}s Vasco-Euskal Herriko Unibertsitatea, 2023.

\bibitem{vos2022robust}
D.~Vos and S.~Verwer, ``Robust optimal classification trees against adversarial examples,'' in {\em AAAI Conference}, vol.~36, pp.~8520--8528, 2022.

\bibitem{dan2020sharp}
C.~Dan, Y.~Wei, and P.~Ravikumar, ``Sharp statistical guarantees for adversarially robust {G}aussian classification,'' in {\em ICML}, pp.~2345--2355, PMLR, 2020.

\bibitem{kumar2021evolutionary}
N.~Kumar, S.~Vimal, K.~Kayathwal, and G.~Dhama, ``Evolutionary adversarial attacks on payment systems,'' in {\em IEEE ICMLA}, pp.~813--818, IEEE, 2021.

\bibitem{cartella2021adversarial}
F.~Cartella {\em et~al.}, ``Adversarial attacks for tabular data: Application to fraud detection and imbalanced data,'' {\em arXiv:2101.08030}, 2021.

\bibitem{kovtun2023hiding}
E.~Kovtun, A.~Ermilova, D.~Berestnev, and A.~Zaytsev, ``Hiding backdoors within event sequence data via poisoning attacks,'' {\em arXiv:2308.10201}, 2023.

\bibitem{carlens2023state}
H.~Carlens, ``State of competitive machine learning in 2022,'' {\em ML Contests Research}, 2023.
\newblock https://mlcontests.com/state-of-competitive-data-science-2022.

\bibitem{dong2023competition}
Y.~Dong, C.~Liu, W.~Xiang, H.~Su, and J.~Zhu, ``Competition on robust deep learning,'' {\em National Science Review}, vol.~10, no.~6, p.~nwad087, 2023.

\bibitem{curry2007detecting}
C.~Curry, R.~L. Grossman, D.~Locke, S.~Vejcik, and J.~Bugajski, ``Detecting changes in large data sets of payment card data: a case study,'' in {\em ACM SIGKDD}, pp.~1018--1022, 2007.

\bibitem{chung2014empirical}
J.~Chung, C.~Gulcehre, K.~Cho, and Y.~Bengio, ``Empirical evaluation of gated recurrent neural networks on sequence modeling,'' {\em arXiv:1412.3555}, 2014.

\bibitem{lecuyer2019certified}
M.~Lecuyer, V.~Atlidakis, R.~Geambasu, D.~Hsu, and S.~Jana, ``Certified robustness to adversarial examples with differential privacy,'' in {\em IEEE symposium on security and privacy (SP)}, pp.~656--672, IEEE, 2019.

\bibitem{prokhorenkova2018catboost}
L.~Prokhorenkova, G.~Gusev, A.~Vorobev, A.~V. Dorogush, and A.~Gulin, ``Catboost: unbiased boosting with categorical features,'' {\em NeurIPS}, vol.~31, 2018.

\bibitem{ke2017lightgbm}
G.~Ke {\em et~al.}, ``{LightGBM}: A highly efficient gradient boosting decision tree,'' {\em NeurIPS}, vol.~30, 2017.

\bibitem{chen2019robust}
H.~Chen, H.~Zhang, D.~Boning, and C.-J. Hsieh, ``Robust decision trees against adversarial examples,'' in {\em ICML}, pp.~1122--1131, PMLR, 2019.

\bibitem{calzavara2019adversarial}
S.~Calzavara, C.~Lucchese, and G.~Tolomei, ``Adversarial training of gradient-boosted decision trees,'' in {\em CIKM}, pp.~2429--2432, 2019.

\bibitem{madry2018towards}
A.~Madry, A.~Makelov, L.~Schmidt, D.~Tsipras, and A.~Vladu, ``Towards deep learning models resistant to adversarial attacks,'' in {\em ICLR}, 2018.

\bibitem{wong2019fast}
E.~Wong, L.~Rice, and J.~Z. Kolter, ``Fast is better than free: Revisiting adversarial training,'' in {\em ICLR}, 2019.

\bibitem{liu2020adversarial}
X.~Liu {\em et~al.}, ``Adversarial training for large neural language models,'' {\em arXiv:2004.08994}, 2020.

\bibitem{shen2018deep}
G.~Shen, Q.~Tan, H.~Zhang, P.~Zeng, and J.~Xu, ``Deep learning with gated recurrent unit networks for financial sequence predictions,'' {\em Procedia computer science}, vol.~131, pp.~895--903, 2018.

\bibitem{mei2017neural}
H.~Mei and J.~M. Eisner, ``The neural hawkes process: A neurally self-modulating multivariate point process,'' {\em NeurIPS}, vol.~30, 2017.

\bibitem{loshchilov2018decoupled}
I.~Loshchilov and F.~Hutter, ``Decoupled weight decay regularization,'' in {\em ICLR}, 2018.

\end{thebibliography}


\begin{thebibliography}{10}

\bibitem{ala2022deep}
M.~Ala’raj, M.~F. Abbod, M.~Majdalawieh, and L.~Jum’a.
\newblock A deep learning model for behavioural credit scoring in banks.
\newblock {\em Neural Computing and Applications}, pages 1--28, 2022.

\bibitem{babaev2022coles}
D.~Babaev, N.~Ovsov, I.~Kireev, M.~Ivanova, G.~Gusev, I.~Nazarov, and A.~Tuzhilin.
\newblock Co{LES}: contrastive learning for event sequences with self-supervision.
\newblock In {\em Proceedings of the 2022 International Conference on Management of Data}, pages 1190--1199, 2022.

\bibitem{babaev2019rnn}
D.~Babaev, M.~Savchenko, A.~Tuzhilin, and D.~Umerenkov.
\newblock {ET-RNN}: Applying deep learning to credit loan applications.
\newblock In {\em ACM SIGKDD}, 2019.

\bibitem{begicheva2021bank}
M.~Begicheva and A.~Zaytsev.
\newblock Bank transactions embeddings help to uncover current macroeconomics.
\newblock In {\em 2021 20th IEEE International Conference on Machine Learning and Applications (ICMLA)}, pages 1742--1748. IEEE, 2021.

\bibitem{calzavara2019adversarial}
S.~Calzavara, C.~Lucchese, and G.~Tolomei.
\newblock Adversarial training of gradient-boosted decision trees.
\newblock In {\em Proceedings of the 28th ACM international conference on information and knowledge management}, pages 2429--2432, 2019.

\bibitem{carlens2023state}
H.~Carlens.
\newblock State of competitive machine learning in 2022.
\newblock {\em ML Contests Research}, 2023.
\newblock https://mlcontests.com/state-of-competitive-data-science-2022.

\bibitem{cartella2021adversarial}
F.~Cartella, O.~Anunciacao, Y.~Funabiki, D.~Yamaguchi, T.~Akishita, and O.~Elshocht.
\newblock Adversarial attacks for tabular data: Application to fraud detection and imbalanced data.
\newblock {\em arXiv preprint arXiv:2101.08030}, 2021.

\bibitem{chen2019robust}
H.~Chen, H.~Zhang, D.~Boning, and C.-J. Hsieh.
\newblock Robust decision trees against adversarial examples.
\newblock In {\em International Conference on Machine Learning}, pages 1122--1131. PMLR, 2019.

\bibitem{chung2014empirical}
J.~Chung, C.~Gulcehre, K.~Cho, and Y.~Bengio.
\newblock Empirical evaluation of gated recurrent neural networks on sequence modeling.
\newblock {\em arXiv preprint arXiv:1412.3555}, 2014.

\bibitem{croce2021robustbench}
F.~Croce, M.~Andriushchenko, V.~Sehwag, E.~Debenedetti, N.~Flammarion, M.~Chiang, P.~Mittal, and M.~Hein.
\newblock Robustbench: a standardized adversarial robustness benchmark.
\newblock In {\em Thirty-fifth Conference on Neural Information Processing Systems Datasets and Benchmarks Track (Round 2)}, 2021.

\bibitem{curry2007detecting}
C.~Curry, R.~L. Grossman, D.~Locke, S.~Vejcik, and J.~Bugajski.
\newblock Detecting changes in large data sets of payment card data: a case study.
\newblock In {\em Proceedings of the 13th ACM SIGKDD international conference on Knowledge discovery and data mining}, pages 1018--1022, 2007.

\bibitem{dan2020sharp}
C.~Dan, Y.~Wei, and P.~Ravikumar.
\newblock Sharp statistical guarantees for adversarially robust {G}aussian classification.
\newblock In {\em International Conference on Machine Learning}, pages 2345--2355. PMLR, 2020.

\bibitem{dixon2020machine}
M.~F. Dixon, I.~Halperin, and P.~Bilokon.
\newblock {\em Machine learning in finance}, volume 1170.
\newblock Springer, 2020.

\bibitem{dong2020benchmarking}
Y.~Dong, Q.-A. Fu, X.~Yang, T.~Pang, H.~Su, Z.~Xiao, and J.~Zhu.
\newblock Benchmarking adversarial robustness on image classification.
\newblock In {\em proceedings of the IEEE/CVF conference on computer vision and pattern recognition}, pages 321--331, 2020.

\bibitem{dong2023competition}
Y.~Dong, C.~Liu, W.~Xiang, H.~Su, and J.~Zhu.
\newblock Competition on robust deep learning.
\newblock {\em National Science Review}, 10(6):nwad087, 2023.

\bibitem{fursov2021adversarial}
I.~Fursov, M.~Morozov, N.~Kaploukhaya, E.~Kovtun, R.~Rivera-Castro, G.~Gusev, D.~Babaev, I.~Kireev, A.~Zaytsev, and E.~Burnaev.
\newblock Adversarial attacks on deep models for financial transaction records.
\newblock In {\em ACM SIGKDD}, 2021.

\bibitem{goodfellow2013challenges}
I.~J. Goodfellow, D.~Erhan, P.~L. Carrier, A.~Courville, M.~Mirza, B.~Hamner, W.~Cukierski, Y.~Tang, D.~Thaler, D.-H. Lee, et~al.
\newblock Challenges in representation learning: A report on three machine learning contests.
\newblock In {\em Neural Information Processing: 20th International Conference, ICONIP 2013, Daegu, Korea, November 3-7, 2013. Proceedings, Part III 20}, pages 117--124. Springer, 2013.

\bibitem{jueguen2023broadening}
J.~V. Jueguen.
\newblock {\em Broadening the Horizon of Adversarial Attacks in Deep Learning}.
\newblock PhD thesis, Universidad del Pa{\'\i}s Vasco-Euskal Herriko Unibertsitatea, 2023.

\bibitem{ke2017lightgbm}
G.~Ke, Q.~Meng, T.~Finley, T.~Wang, W.~Chen, W.~Ma, Q.~Ye, and T.-Y. Liu.
\newblock Lightgbm: A highly efficient gradient boosting decision tree.
\newblock {\em Advances in neural information processing systems}, 30, 2017.

\bibitem{khorshidi2022adversarial}
S.~Khorshidi, B.~Wang, and G.~Mohler.
\newblock Adversarial attacks on deep temporal point process.
\newblock In {\em 2022 21st IEEE International Conference on Machine Learning and Applications (ICMLA)}, pages 1--8. IEEE, 2022.

\bibitem{kovtun2023hiding}
E.~Kovtun, A.~Ermilova, D.~Berestnev, and A.~Zaytsev.
\newblock Hiding backdoors within event sequence data via poisoning attacks.
\newblock {\em arXiv preprint arXiv:2308.10201}, 2023.

\bibitem{kumar2021evolutionary}
N.~Kumar, S.~Vimal, K.~Kayathwal, and G.~Dhama.
\newblock Evolutionary adversarial attacks on payment systems.
\newblock In {\em 2021 20th IEEE International Conference on Machine Learning and Applications (ICMLA)}, pages 813--818. IEEE, 2021.

\bibitem{lecuyer2019certified}
M.~Lecuyer, V.~Atlidakis, R.~Geambasu, D.~Hsu, and S.~Jana.
\newblock Certified robustness to adversarial examples with differential privacy.
\newblock In {\em 2019 IEEE symposium on security and privacy (SP)}, pages 656--672. IEEE, 2019.

\bibitem{loshchilov2018decoupled}
I.~Loshchilov and F.~Hutter.
\newblock Decoupled weight decay regularization.
\newblock In {\em International Conference on Learning Representations}, 2018.

\bibitem{mei2017neural}
H.~Mei and J.~M. Eisner.
\newblock The neural hawkes process: A neurally self-modulating multivariate point process.
\newblock {\em Advances in neural information processing systems}, 30, 2017.

\bibitem{munoz2023combination}
R.~Mu{\~n}oz-Cancino, C.~Bravo, S.~A. R{\'\i}os, and M.~Gra{\~n}a.
\newblock On the combination of graph data for assessing thin-file borrowers’ creditworthiness.
\newblock {\em Expert Systems with Applications}, 213:118809, 2023.

\bibitem{prokhorenkova2018catboost}
L.~Prokhorenkova, G.~Gusev, A.~Vorobev, A.~V. Dorogush, and A.~Gulin.
\newblock Catboost: unbiased boosting with categorical features.
\newblock {\em Advances in neural information processing systems}, 31, 2018.

\bibitem{qiu2022adversarial}
S.~Qiu, Q.~Liu, S.~Zhou, and W.~Huang.
\newblock Adversarial attack and defense technologies in natural language processing: A survey.
\newblock {\em Neurocomputing}, 492:278--307, 2022.

\bibitem{shchur2021neural}
O.~Shchur, A.~C. T{\"u}rkmen, T.~Januschowski, and S.~G{\"u}nnemann.
\newblock Neural temporal point processes: A review.
\newblock {\em arXiv preprint arXiv:2104.03528}, 2021.

\bibitem{shen2018deep}
G.~Shen, Q.~Tan, H.~Zhang, P.~Zeng, and J.~Xu.
\newblock Deep learning with gated recurrent unit networks for financial sequence predictions.
\newblock {\em Procedia computer science}, 131:895--903, 2018.

\bibitem{sukharev2020ews}
I.~Sukharev, V.~Shumovskaia, K.~Fedyanin, M.~Panov, and D.~Berestnev.
\newblock Ews-gcn: Edge weight-shared graph convolutional network for transactional banking data.
\newblock In {\em 2020 IEEE International Conference on Data Mining (ICDM)}, pages 1268--1273. IEEE, 2020.

\bibitem{szegedy2014intriguing}
C.~Szegedy, W.~Zaremba, I.~Sutskever, J.~Bruna, D.~Erhan, I.~Goodfellow, and R.~Fergus.
\newblock Intriguing properties of neural networks.
\newblock In {\em 2nd International Conference on Learning Representations, ICLR 2014}, 2014.

\bibitem{tabassi2019taxonomy}
E.~Tabassi, K.~J. Burns, M.~Hadjimichael, A.~D. Molina-Markham, and J.~T. Sexton.
\newblock A taxonomy and terminology of adversarial machine learning.
\newblock {\em NIST IR}, 2019:1--29, 2019.

\bibitem{tripathi2012survey}
K.~K. Tripathi and M.~A. Pavaskar.
\newblock Survey on credit card fraud detection methods.
\newblock {\em International Journal of Emerging Technology and Advanced Engineering}, 2(11):721--726, 2012.

\bibitem{vos2022robust}
D.~Vos and S.~Verwer.
\newblock Robust optimal classification trees against adversarial examples.
\newblock In {\em Proceedings of the AAAI Conference on Artificial Intelligence}, volume~36, pages 8520--8528, 2022.

\bibitem{wang2022deep}
C.~Wang and Z.~Xiao.
\newblock A deep learning approach for credit scoring using feature embedded transformer.
\newblock {\em Applied Sciences}, 12(21):10995, 2022.

\bibitem{xu2020adversarial}
H.~Xu, Y.~Ma, H.-C. Liu, D.~Deb, H.~Liu, J.-L. Tang, and A.~K. Jain.
\newblock Adversarial attacks and defenses in images, graphs and text: A review.
\newblock {\em International Journal of Automation and Computing}, 17:151--178, 2020.

\bibitem{zhuzhel2023continuous}
V.~Zhuzhel, V.~Grabar, G.~Boeva, A.~Zabolotnyi, A.~Stepikin, V.~Zholobov, M.~Ivanova, M.~Orlov, I.~Kireev, E.~Burnaev, et~al.
\newblock Continuous-time convolutions model of event sequences.
\newblock {\em arXiv preprint arXiv:2302.06247}, 2023.

\end{thebibliography}

% \clearpage
\appendix
% \section{Appendix}

In the appendix, we provide additional technical details on the methods used, a deeper analysis of the competition results, and additional experiments.

\subsection{Dataset separation}
\label{sec:data_separation}
\begin{figure}[t!]
    \centering
    \includegraphics[width=0.99\columnwidth]{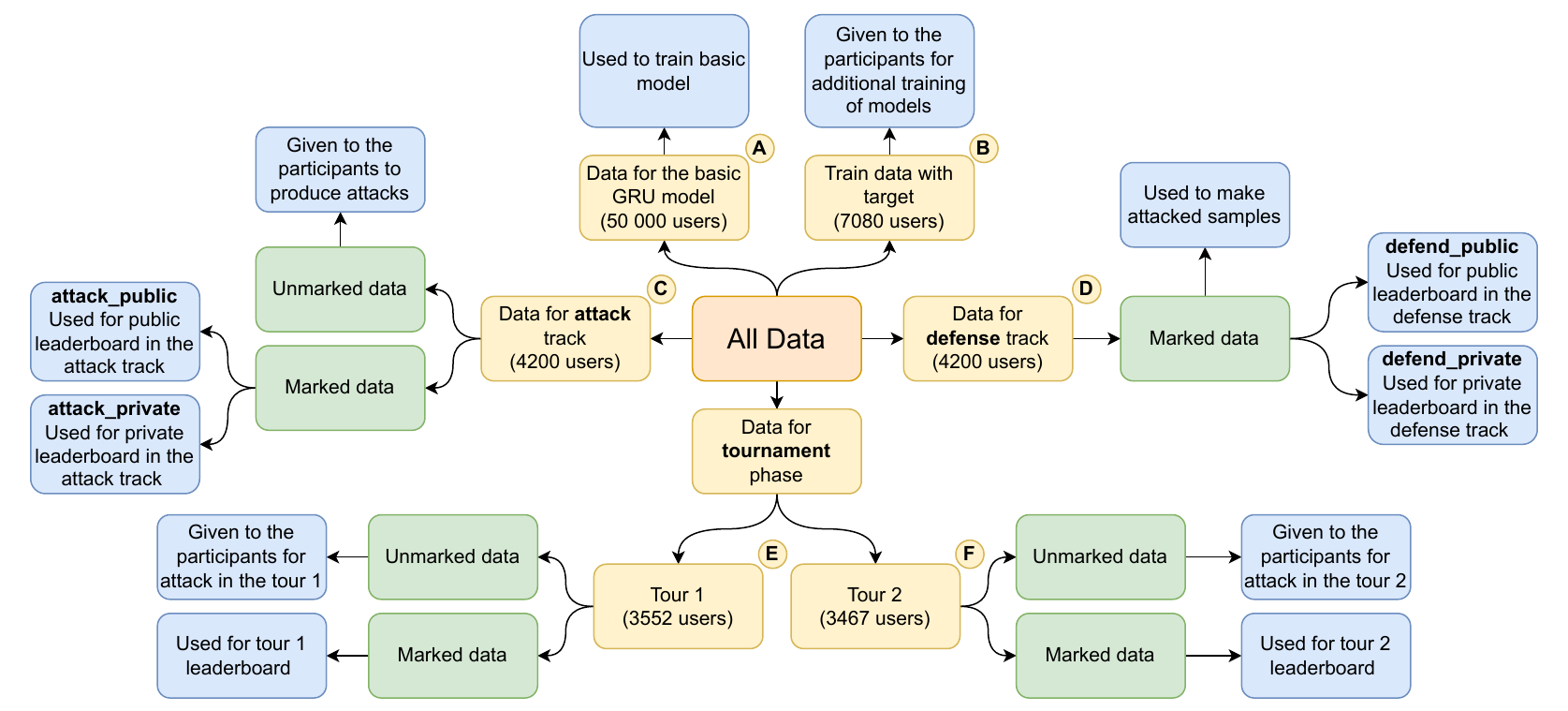}
    \caption{Data split structure}
    \label{fig:data_sep}
\end{figure}

We have numerous phases of learning and evaluation. 
To prevent data leakage, each stage uses its own data fold. 
This leads us to the complex data separation structure presented in the picture \ref{fig:data_sep} and discussed below.

% A
The first data fold indicated by the letter \textbf{A} in the figure is the data for learning a basic GRU model discussed in the next section. 
It contains transactions from $50000$ users hidden from the participants.
%выкладываем ли мы эти данные?
% B
The second data fold \textbf{B} was used for the finetuning of models and was available for participants in the pre-tournament phase in both attack and defense tracks. This fold is publicly available and contains transactions from 7080 users with marking.
% C
Fold \textbf{C} contains data from 4200 users and was used for the attack pre-tournament stage. As at this track participants have to provide attacked sequences, so only unmarked sequences were posted publicly. The marking of these data was used in the public and private leaderboards to compute metrics.
% D 
Fold \textbf{D} is the part of data for the defense track of the pre-tournament stage. This data was not publicly available for participants and was used to produce transaction sequences attacked by the baseline attack and to calculate metrics for public and private leaderboards.
% E & F
Lastly, folds \textbf{E} and \textbf{F} were used in the tour 1 and tour 2 of the tournament. These folds contain data from 3552 and 3467 users respectively. Unmarked data were provided to the participants for the attack track to prepare attacks, and marking was used in leaderboard compilation for these stages.

\subsection{Technical details of the best performing methods and baselines}

\subsubsection{Training of a model for the attack}
\label{sec:base_model}

\begin{figure}[h!]
    \centering
    \includegraphics[width=0.99\columnwidth]{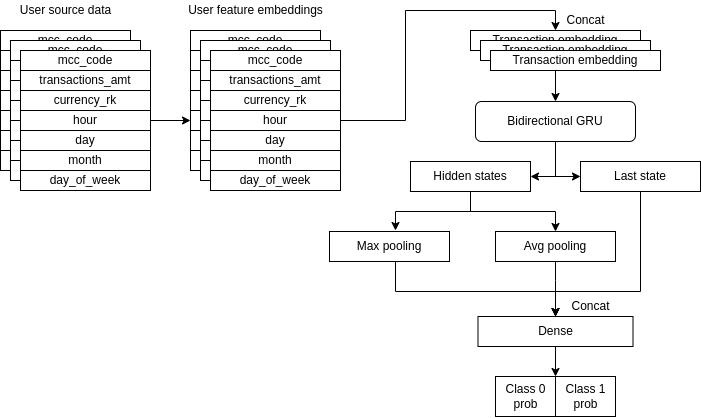}
    \caption{Scheme of the attacked GRU model}
    \label{fig:gru_model_scheme}
\end{figure}

To train the model for the attack, we utilize transactions from $50000$ customers. 
On top of preprocessed data, we train a GRU neural network suitable for financial problems and event sequences in general~\cite{shen2018deep,mei2017neural} and for processing financial transactions~\cite{babaev2019rnn}.
The scheme of the model that includes preprocessing is available in Figure~\ref{fig:gru_model_scheme}.
After the training, we obtained a model with a $0.7$ ROC AUC value, which is typical for the considered target.
Below we provide additional details on the training process.

For each transaction, we obtain embeddings, where each feature has a separate embedding vector. 
These embeddings are concatenated, thereby representing each transaction as a single vector. A sample for each customer consists of $300$ transactions, with the output is the predicted default class probability.

Preprocessing includes feature generation and transformation.
Each transaction was enriched by adding categorical temporal features such as the hour, day, day of the week, and month. 
The transaction amount was binned into 100 quantiles, transforming all features into categorical variables.

We employ the AdamW optimizer~\cite{loshchilov2018decoupled} with a learning rate set at $3e-4$, using the binary cross entropy as the loss function. 
The training adopts two regularization techniques: a spatial dropout with a probability of $0.5$ before the GRU layer and a dropout layer with the same probability before the linear layer.

The model code and its weights after training can be found on the \href{https://vorsineo.github.io/adv_ml_tournament/#subsection4-1}{website}.

\subsubsection{The best attack}

% make experiments to validate the approach???
% напрашивается ablation study

% \begin{figure*}[h!]
%     \centering
%     \begin{subfigure}[t]{0.35\textwidth}
%         \centering
%         \includegraphics[width=\textwidth]{figures/attack_shake.pdf}
%         \caption{Attack}
%         \label{fig:attack_shakeup}
%     \end{subfigure}%
%     ~ 
%     \begin{subfigure}[t]{0.35\textwidth}
%         \centering
%         \includegraphics[width=\textwidth]{figures/defense_shake.pdf}
%         \caption{Defense}
%         \label{fig:defense_shakeup}
%     \end{subfigure}
%     \caption{Difference between private and public scores for the attack \ref{fig:attack_shakeup} and defense \ref{fig:defense_shakeup} track of the competition}
% \end{figure*}

The best attack approach is a variation of SamplingFool~\cite{fursov2021adversarial} that showed the best results for financial transaction data.
In particular, the attack imitates a random search over a discrete space of sequences of transactions:
\begin{enumerate}
    \item At each iteration of an attack we generate $k$ candidates and rank them according to a model $\hat{f}(\mathbf{x})$.
    \item We select top $k_0$ changes and move to the next iteration.
\end{enumerate}
As the number of iterations is equal to the admissible number of changes, in the end, we have an admissible set of changes.

\paragraph*{Hyperparameters}
The attack uses $k = 10 000$ candidates at each step and $k_0 = 100$.
As a model $\hat{f}(\mathbf{x})$, we use a given GRU-based model or a set of gradient boosting models.
The use of a given model or an ensemble model was randomly selected with a probability of $0.5$ to make possible attacks on diverse models.

% why we can't use the model scores itself?
The ensemble for the imitation of the score of the true model is $100$ of gradient boosting models that use MCC codes and amount (so, they are features an attacker can affect).
We use a CatBoost gradient boosting implementation.
The derivative features are a common set of aggregates for financial transactions.
To diversify the models and improve their quality, we learn each model using a subset of generated aggregates.
Each separate model was significantly worse than the baseline model, but in total, the quality of models doesn't affect the quality of attacks.

\begin{figure*}[t!]
    \centering
    \centering
    \begin{subfigure}[t]{0.21\textwidth}
        \centering
        \includegraphics[width=\textwidth]{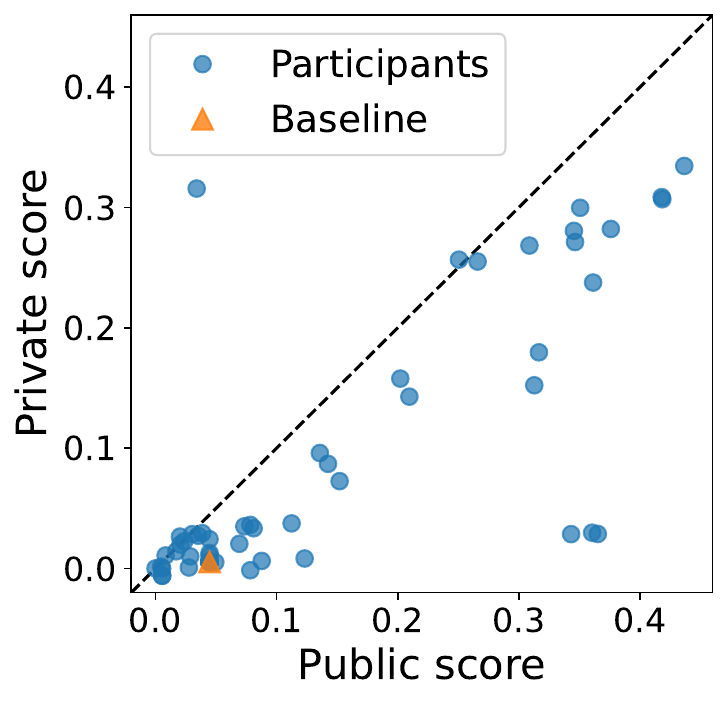}
        \caption{Attack differences}
        \label{fig:attack_shakeup}
    \end{subfigure}%
    ~ 
    \begin{subfigure}[t]{0.22\textwidth}
        \centering
        \includegraphics[width=\textwidth]{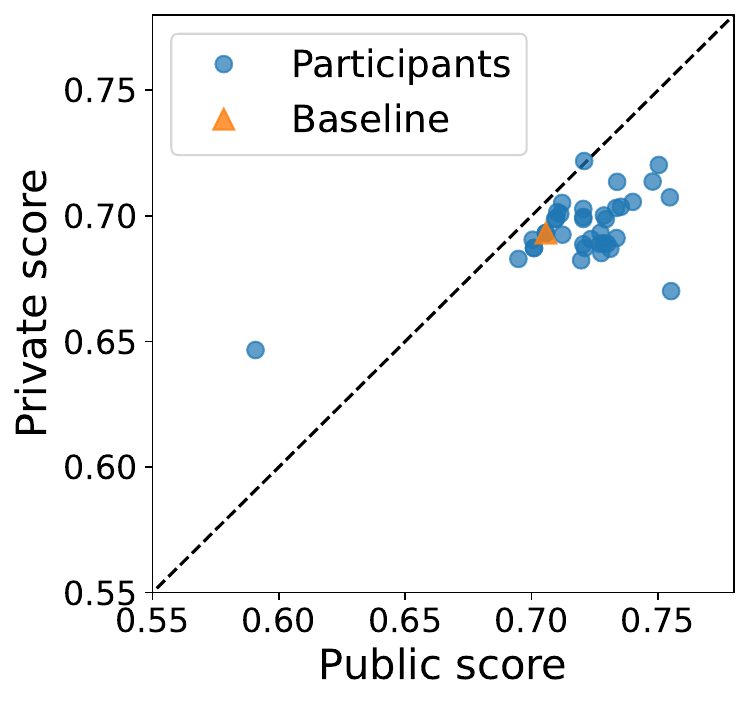}
        \caption{Defense differences}
        \label{fig:defense_shakeup}
    \end{subfigure}
    ~
    \begin{subfigure}[t]{0.255\textwidth}
        \centering
        \includegraphics[width=\textwidth]{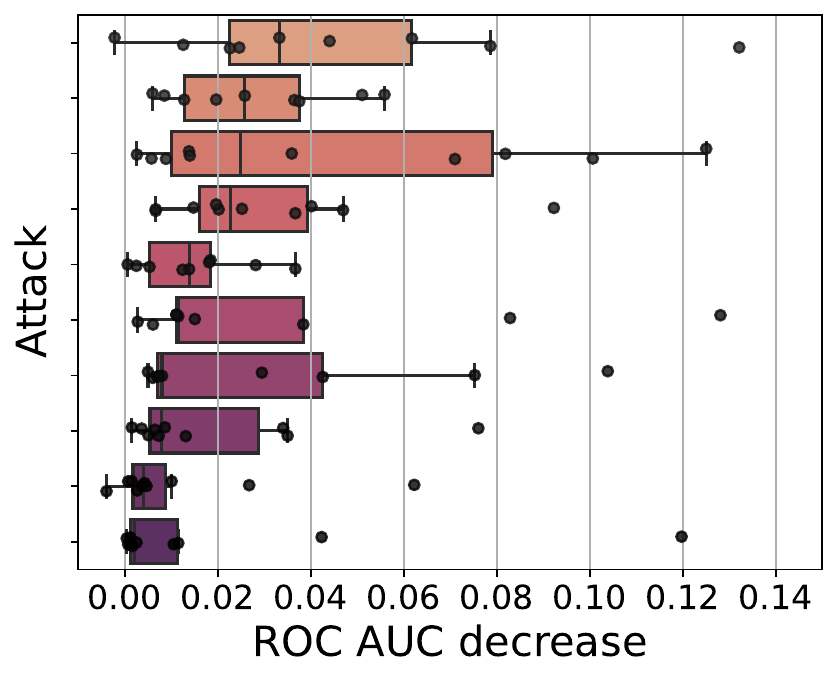}
        \caption{Attack performances}
        \label{fig:comp_attack}
    \end{subfigure}%
    ~ 
    \begin{subfigure}[t]{0.255\textwidth}
        \centering
        \includegraphics[width=\textwidth]{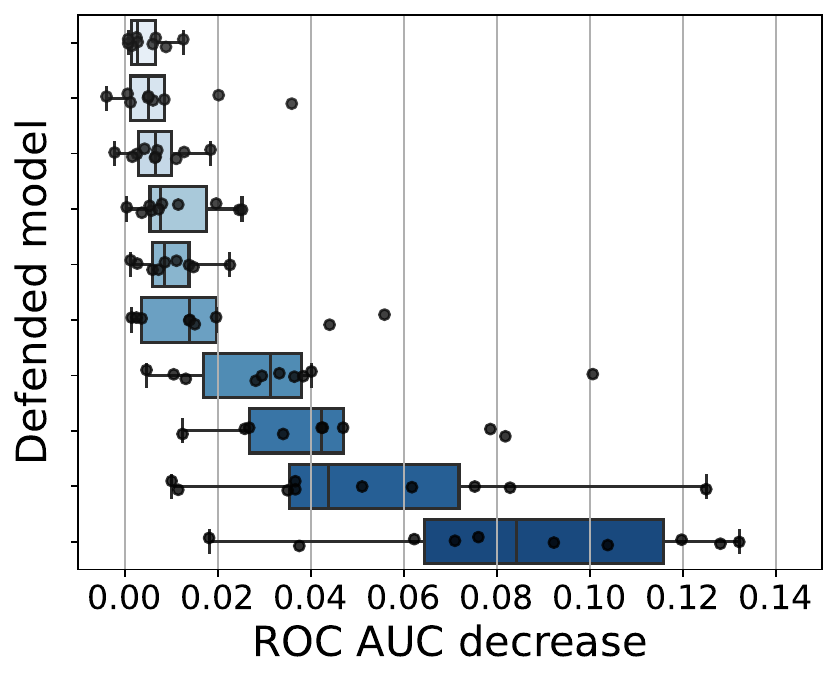}
        \caption{Defense performances}
        \label{fig:comp_defense}
    \end{subfigure}
    \caption{On the left pictures \ref{fig:attack_shakeup} and  \ref{fig:defense_shakeup} difference between private and public scores for the attack \ref{fig:attack_shakeup} and defense \ref{fig:defense_shakeup} track of the competition is presented. On the right pictures \ref{fig:comp_attack} and \ref{fig:comp_defense} performance of the attacks \ref{fig:comp_attack} and defended models \ref{fig:comp_defense} during the black box tournament stage is presented. Each row represents a single attack / defended model. Each dot represents a single attacked model / performed attack. Better to view in zoom.}
\end{figure*}

\subsubsection{The best defence}
We note that the permutation of transactions doesn't affect the model score.
So, to make the model more robust, we can perform a fixed number of permutations and get predictions for the initial model.
Averaging these predictions gives the most robust option.
So, the defense has two components: a permutation algorithm and a model used to evaluate permuted sequences of transactions.

The defense model uses a distilled model from the base one.
It is also an ensemble that was learned to distil the big model using only a subset of given features.
We trained here another gradient boosting model.

\subsection{Additional results}

\subsubsection{Analysis of the competitions' results}
Figure~\ref{fig:attack_shakeup} presents a comparison of public and private scores for participants during the attack stage of the competition.
Due to overfitting, the private scores are almost always lower than the public score, making the attack less powerful.
We suspect that a single point with the reverse effect is a result of the purposeful masking of the public score by a participant. 

Figure~\ref{fig:defense_shakeup} presents a comparison of public and private scores for participants during the defense stage of the competition.
The results suggest significant differences between public and private scores, with the latter not being available to participants during the competition, suggesting another type of overfitting.

\subsubsection{Analysis of competitions tournament}
\label{sec:roc_auc_decrease}
Figures~\ref{fig:comp_attack} and \ref{fig:comp_defense} present the performance of all attacks against all defenses.
One can see in more detail the difference in performance between top-10 attacks and defenses.

\subsubsection{Sensitivity to the number of changed transactions}
\begin{figure}[h]
    \centering
    \includegraphics[width=0.55\columnwidth]{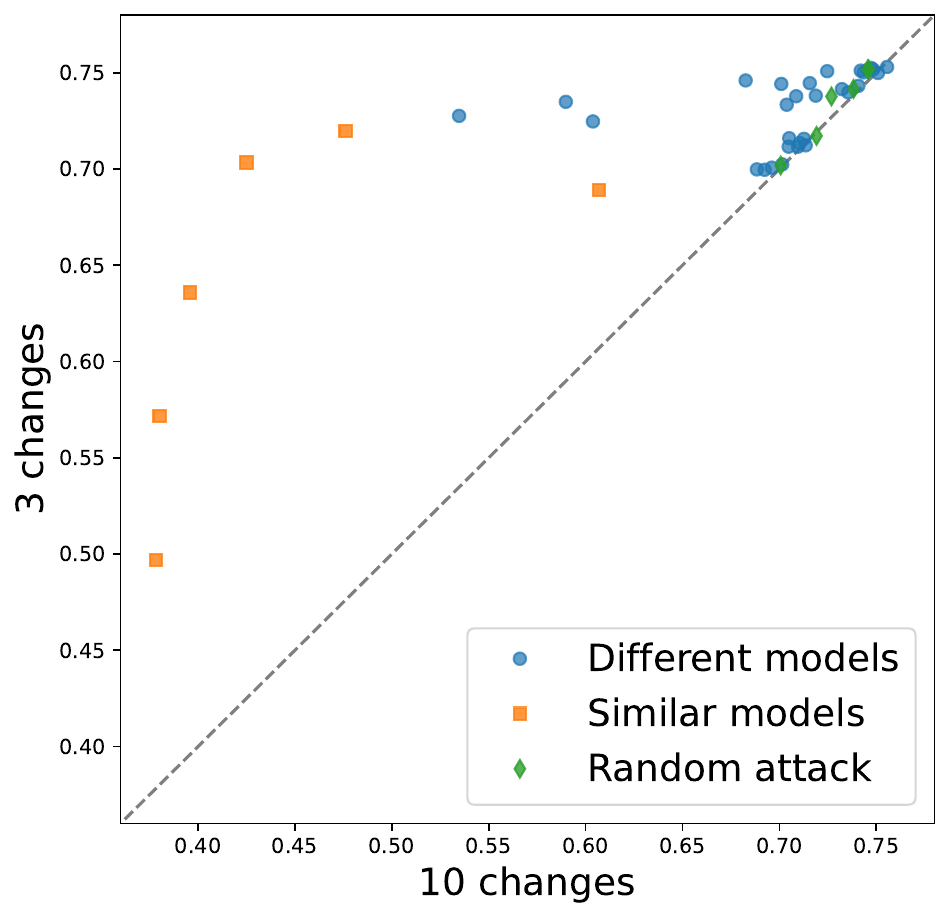}
    \caption{Comparison of ROC AUC values after attacks for $3$ and $10$ admissible changes: models of similar architecture, of different architecture, and with random changes.}
    \label{fig:number_of_changes_plot}
\end{figure}

To make a comparison, we varied various constraints for attack and found the most significant number of possible changes $k$.
We present the comparison for a different number of possible changes in Figure~\ref{fig:number_of_changes_plot} %and the Appendix \ref{sec:ablation}.
If the models are similar, adding more steps to attack would help a lot.
If they are different, the change is little.

\begin{figure}[h]
    \centering
    \includegraphics[width=0.5\textwidth]{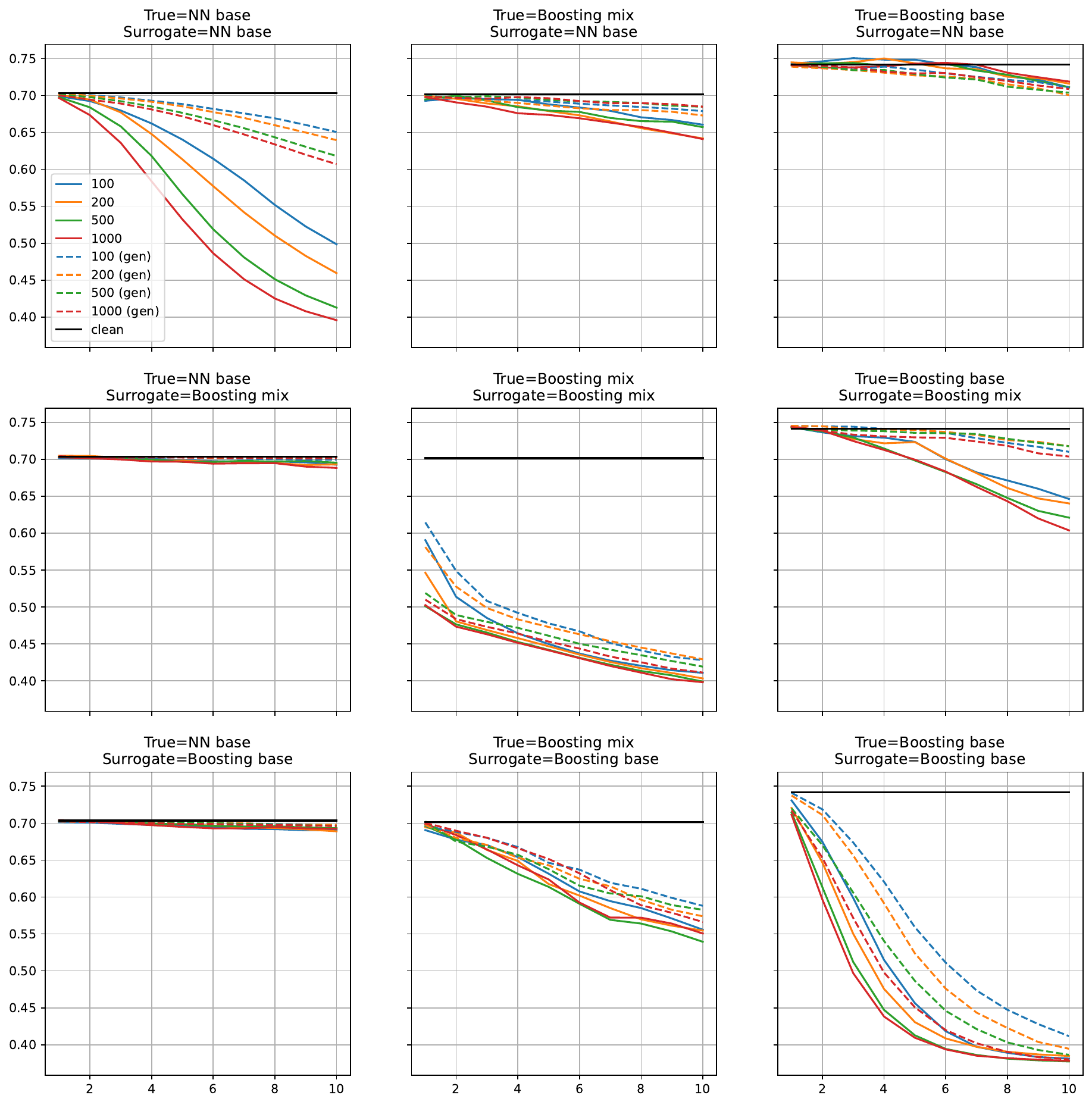}
    \caption{ROC AUC versus the number of changes for different types of True and Surrogate (used for attack) models. The X-axis is the number of possible substitutions during an attack, and the Y-axis is the ROC AUC metric for an attacked model after an attack. Better to view in zoom.}
    \label{fig:number_of_changes_full_plot}
\end{figure}

\subsubsection{Sensitivity and ablation studies}
\label{sec:ablation}
We conduct additional experiments to examine how the attack's quality is affected by the choice of model architecture.
Other considered changes are in the number of admissible changes $k$ and the number of options tried during each step.
Also, we consider an alternative that generates more realistic and concealed substitutions. 
It allows changes only of less prominent MCCs and uses more realistic amounts for MCCs suggested by~\cite{fursov2021adversarial}. 
We use the prefix \textbf{gen} to mark such approaches.

The results are presented in the Figure~\ref{fig:number_of_changes_full_plot}.
We see a continuous trend of decreasing model performance as we allow an attack to change more tokens in a sequence.
If an attacked surrogate model and a true model are close to each other, the difference in performance is drastic, reaching $0.4$ ROC AUC for $10$ possible substitutions.
If models are close to each other, e.g., a single Boosting model and a Boosting mix ensemble, the attack also works, but the results are weaker.
On the other hand, if we use different architectures for different models, we observe performance drops close to that related to random changes.
With respect to the parameter $k$, we observe that generating more plausible selection options at a single iteration leads to stronger attacks, even in the case of significant differences between true and attacked models.

\subsubsection{Results for Churn dataset}
\label{sec:res_churn}

% \paragraph{Churn}
The Churn dataset was taken from a competition \footnote{https://boosters.pro/championship/rosbank1/overview} and used previously in \cite{babaev2022coles}. 
It contains the same features of bank clients' transactions as the Default dataset, but the number of MCC categories is $\approx 350$. 
The target is whether the client will leave the bank or not, and the number of positive and negative labels in this task is almost equal.
The length of sequences ranges from 1 to 784 and averages 100.
Table~\ref{table:churn_results} presents results for the \emph{Churn} dataset.
\begin{table}[h]
      \centering
      % \footnotesize
        \begin{tabular}{lccccc}
\hline
Attack & No & \multicolumn{2}{c}{NN}  & \multicolumn{2}{c}{Boosting}  \\
& attack & \multicolumn{2}{c}{base} & \multicolumn{2}{c}{base}\\
\hline
Change \# &        & 3    & 5 & 3    & 5 \\
\hline
NN base & 0.767 & \underline{0.369} & \underline{0.335} & 0.719 & 0.712 \\
NN mix & 0.764 & 0.667 & \underline{0.593} & 0.732 & 0.718 \\
Boost. base & 0.823 & 0.786 & 0.764 & \underline{0.388} & \underline{0.338} \\
Boost. mix & 0.800 & 0.795 & 0.794 & 0.751 & 0.715 \\
Boost. mix filter & 0.799 & 0.798 & 0.797 & 0.798 & 0.776 \\
\hline
\end{tabular}
\captionof{table}{ROC AUC values for various attacks and models for the \emph{Churn} dataset with different numbers of possible changes. We underline $\leq 0.6$ ROC AUC scores.}
\label{table:churn_results}
\end{table}

\end{document}